\documentclass[10pt,journal,compsoc]{IEEEtran}

\usepackage[utf8]{inputenc} 
\usepackage[T1]{fontenc}    
\usepackage{url}            
\usepackage{booktabs}       
\usepackage{amsfonts}       
\usepackage{amsmath}
\usepackage{nicefrac}       
\usepackage{microtype}      
\usepackage{amsthm}
\usepackage{amssymb}
\usepackage{rotating}
\usepackage{booktabs}
\usepackage{graphicx}
\usepackage{subfigure}
\usepackage{caption}
\usepackage{wrapfig}
\usepackage{bm}
\usepackage{color}
\usepackage{xcolor}
\usepackage{multirow}
\usepackage{mathtools}
\usepackage{lscape}
\usepackage{hyperref}
\usepackage{url}
\usepackage{bbm}
\newcommand{\eat}[1]{}

\usepackage{amsmath,amsfonts,bm}

\def\eqref#1{equation~\ref{#1}}

\def\1{\bm{1}}

\DeclareMathAlphabet{\mathsfit}{\encodingdefault}{\sfdefault}{m}{sl}
\SetMathAlphabet{\mathsfit}{bold}{\encodingdefault}{\sfdefault}{bx}{n}

\definecolor{citeblue}{HTML}{0071bc}
\hypersetup{
  colorlinks,
  linkcolor = citeblue,
  citecolor = citeblue,
}

\usepackage{color,colortbl}
\definecolor{ballblue}{rgb}{0.13, 0.67, 0.80}
\definecolor{amaranth}{rgb}{0.90, 0.17, 0.31}
\definecolor{olive}{rgb}{0.5, 0.5, 0.0}
\definecolor{Gray}{gray}{0.85}
\newcolumntype{a}{>{\columncolor{Gray}}c}
\definecolor{blush}{rgb}{0.87, 0.36, 0.51}
\newcommand{\etal}{\textit{et al.}}

\ifCLASSOPTIONcompsoc
  \usepackage[nocompress]{cite}
\else
  \usepackage{cite}
\fi

\ifCLASSINFOpdf
\else
\fi

\begin{document}

\title{A Survey on Graph Neural Networks and Graph Transformers in Computer Vision: \\ A Task-Oriented Perspective}

\author{Chaoqi Chen, Yushuang Wu, Qiyuan Dai, Hong-Yu Zhou, Mutian Xu,\\ Sibei Yang, Xiaoguang Han, and Yizhou Yu,~\IEEEmembership{Fellow,~IEEE}
	\IEEEcompsocitemizethanks{
		\IEEEcompsocthanksitem C. Chen, H.-Y. Zhou, and Y. Yu are with the Department of Computer Science, The University of Hong Kong, Hong Kong. (email: cqchen1994@gmail.com, whuzhouhongyu@gmail.com, yizhouy@acm.org).
        \IEEEcompsocthanksitem Y. Wu, M. Xu, and X. Han are with the School of Science and Engineering, The Chinese University of Hong Kong, Shenzhen. Y. Wu and X. Han are also with the Future Network of Intelligence Institute, CUHK-Shenzhen. (email: yushuangwu@link.cuhk.edu.cn, mutianxu@link.cuhk.edu.cn, hanxiaoguang@cuhk.edu.cn).
        \IEEEcompsocthanksitem Q. Dai and S. Yang are with the School of Information Science and Technology, ShanghaiTech University, Shanghai. S. Yang is also with Shanghai Engineering Research Center of Intelligent Vision and Imaging. (email: daiqy2022@shanghaitech.edu.cn, yangsb@shanghaitech.edu.cn).
        \IEEEcompsocthanksitem C. Chen, Y. Wu, Q. Dai, and H.-Y. Zhou contributed equally to this work. Corresponding authors: S. Yang, X. Han, and Y. Yu.
	}}

\markboth{}{Chen \MakeLowercase{\textit{et al.}}: A Survey on Graph Neural Networks and Graph Transformers in Computer Vision}

\IEEEtitleabstractindextext{%
\begin{abstract}
Graph Neural Networks (GNNs) have gained momentum in graph representation learning and boosted the state of the art in a variety of areas, such as data mining (\emph{e.g.,} social network analysis and recommender systems), computer vision (\emph{e.g.,} object detection and point cloud learning), and natural language processing (\emph{e.g.,} relation extraction and sequence learning), to name a few.
With the emergence of Transformers in natural language processing and computer vision,
graph Transformers embed a graph structure into the Transformer architecture to overcome the limitations of local neighborhood aggregation while avoiding strict structural inductive biases.
In this paper, we present a comprehensive review of GNNs and graph Transformers in computer vision from a task-oriented perspective.
Specifically, we divide their applications in computer vision into five categories according to the modality of input data, \emph{i.e.,} 2D natural images, videos, 3D data, vision + language, and medical images.
In each category, we further divide the applications according to a set of vision tasks.
Such a task-oriented taxonomy allows us to examine how each task is tackled by different GNN-based approaches and how well these approaches perform.
Based on the necessary preliminaries,
we provide the definitions and challenges of the tasks, in-depth coverage of the representative approaches, as well as discussions regarding insights, limitations, and future directions.

\end{abstract}

}

\maketitle

\IEEEdisplaynontitleabstractindextext

\IEEEpeerreviewmaketitle
\IEEEraisesectionheading{\section{Introduction}\label{sec:introduction}}
\IEEEPARstart{D}{eep} learning~\cite{lecun2015deep} has brought many breakthroughs to computer vision, where convolutional neural networks (CNN) take a dominant position and become the fundamental infrastructure of many modern vision systems.
In particular, a number of state-of-the-art CNN models, such as AlexNet~\cite{krizhevsky2012imagenet} and ResNet~\cite{he2016deep}, 
have been proposed and achieved unprecedented advances in a variety of vision problems, including image classification, object detection, semantic segmentation, and image processing to name a few. Moreover, existing vision systems take various input modalities as humans do, such as 2D images (\emph{e.g.,} natural and medical images), videos, 3D data (\emph{e.g.,} point clouds and meshes), as well as multimodal inputs (\emph{e.g.,} image + text).

Despite the proliferation of CNN-based methods that excel at dealing with input data defined on regular grids, such as images,
there is an emerging sense in the computer vision community that visual information with an irregular topology is crucial for representation learning but is yet to be thoroughly studied.
Upon observing that human capacity for combinatorial generalization largely relies on their cognitive mechanisms for representing structures and reasoning about relations~\cite{battaglia2018relational},
mimicking human learning and decision-making processes could improve the performance of vision models. 
For instance, in the task of object recognition,
state-of-the-art neural networks prefer to focus on perceiving separate objects,
whereas the dependencies and interactions among different objects have received scant attention.

Moreover, compared to natural graph data, such as social networks and biological protein-protein networks, that has intrinsic edge connections and the notion of nodes,
there is a shortage of principled methods for the construction of graphs (\emph{e.g.,} relation graphs) from regular grid data (\emph{e.g.,} images and temporal signals), and domain knowledge is critical for the success.
On the other hand, certain visual data formats, such as point clouds and meshes,
are naturally not defined on a Cartesian grid and involve sophisticated relational information.
In that sense, both regular and irregular visual data formats would benefit from the exploration of topological structures and relations especially for challenging scenarios, such as understanding complex scenes, learning from limited experiences, and transferring knowledge across domains.


\begin{figure*}[!t]
	\centering
	\includegraphics[width=1\textwidth]{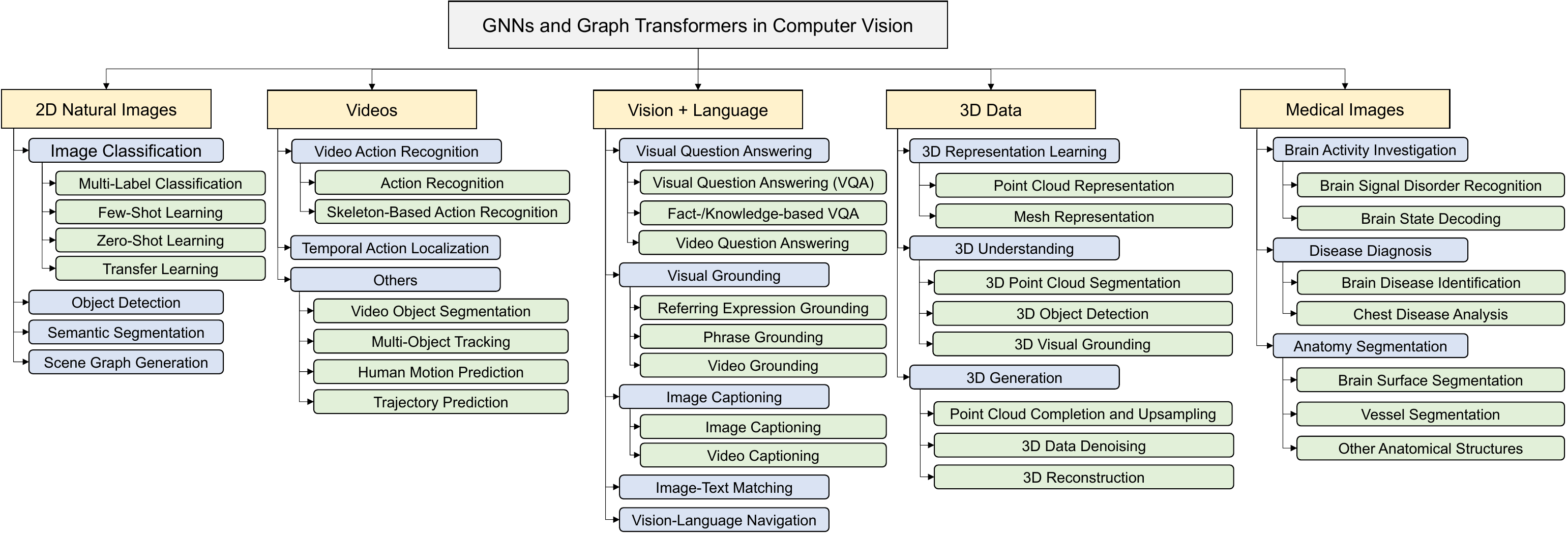}
	\caption{Overview of the landscape of GNNs and graph Transformers in computer vision.}\label{fig1}
\end{figure*}

In the past few years, GNNs~\cite{sperduti1997supervised} have demonstrated ground-breaking performance in modeling graph structures under the umbrella of recent advancements in deep learning.
In the scope of computer vision, much of current GNN-related research has one of the following two objectives: (1) a mixture of GNN and CNN backbones, and (2) a pure GNN architecture for representation learning.
The former typically seeks to improve the long-range modeling ability of CNN-based features and applies to vision tasks that were previously solved using pure CNN architectures, such as image classification and semantic segmentation.
The latter serves as a feature extractor for certain visual data formats, such as point clouds, and was developed in parallel to other approaches. For instance, for the classification of 3D shapes represented as point clouds~\cite{guo2020deep}, there exist three major approaches, namely, pointwise MLP methods, convolution-based methods, and graph-based methods. 

Despite the fruitful advancements, there still does not exist a survey to systematically and timely review how GNN-based computer vision has progressed.
Specifically, we present this literature review as a complete introduction of GNNs in computer vision from a task-oriented perspective, including \emph{(i)} the definitions and challenges of the tasks, \emph{(ii)} in-depth coverage of the representative approaches, and \emph{(iii)} systematic discussions regarding insights and future directions.
In particular, we divide the applications of GNNs in computer vision into five categories according to the modality of input data.
In each category, we further divide the applications according to the computer vision tasks they perform.
We also review graph Transformers used in vision tasks considering their similarity with GNNs in terms of architecture~\cite{dwivedi2020generalization,chen2022structure}.
The organization of this survey is shown in Fig.~\ref{fig1}.
While several surveys (\emph{e.g.,}\cite{krzywda2022graph,pradhyumna2021graph}) have previously reviewed the application of GNNs in certain vision tasks,
we provide a more comprehensive and detailed examination of GNNs and graph Transformers in vision, a better taxonomy of the literature, and present discussions regarding insights, limitations, and potential directions for future research. 

\section{Background and Categorization}
\label{sec:background}

In this section, 
we recap GNNs and graph Transformers used in computer vision. 
Readers could refer to several previous GNN surveys~\cite{wu2020comprehensive,zhou2020graph,rong2020deep} that comprehensively introduce the development of GNNs. 
In addition, we would like to emphasize that many existing GNN-based vision approaches actually use a mixture of CNNs and GNNs, whereas we focus on the GNN side.  

\subsection{Recurrent GNNs}
GNN was initially developed in the form of recurrent GNNs. 
Earlier work~\cite{sperduti1997supervised} in this regime tries to extract node representations from directed acyclic graphs by recurrently using the same set of weights over iterations.  
Scarselli~\etal~\cite{scarselli2008graph} extended such neural networks to process more types of graphs, such as cyclic and undirected graphs. 
They recurrently update the hidden state $\mathbf{h}$ of a node as follows,
\begin{equation}\label{eq:gnn}
        \small
	\mathbf{h}_{i}^{(t+1)} = \sum_{v_j \in \mathcal{N}(v_i)}f \left( \mathbf{x}_{i}, \mathbf{x}_{j}, \mathbf{x}_{ij}^{\mathbf{e}}, \mathbf{h}^{(t)}_{j} \right),
\end{equation}
where $\mathcal{N}(v_i)$ represents the neighborhood of node $v_i$, $f(\cdot)$ is a feedforward neural network, $\mathbf{x}_i \in \mathbb{R}^d$ denotes the features at $v_i$, $\mathbf{x}_{ij}^{\mathbf{e}} \in \mathbb{R}^c$ denotes the features at the edge between $v_i$ and $v_j$, and $t$ is the iteration number.

\subsection{Convolutional GNNs}
\begin{figure}[htb]
	\centering
	\includegraphics[width=0.45\textwidth]{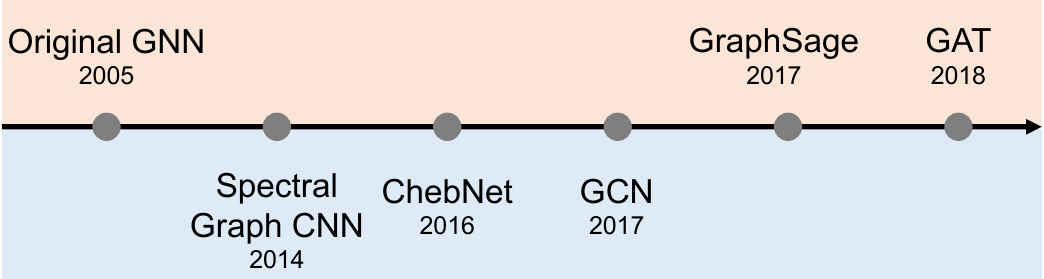}
	\caption{Two types of graph convolutional operations.}\label{fig:GNN-history}
	\vspace{-0.5cm}
\end{figure}

Inspired by the astonishing progress of CNNs in the deep learning era, 
many research efforts have been devoted to generalizing convolution to the graph domain.
Among them, there are two series of approaches (cf. Fig.~\ref{fig:GNN-history}) that garner the most attention in recent years, namely, the spectral approaches~\cite{bruna2014spectral,henaff2015deep,defferrard2016convolutional,kipf2017semi,levie2018cayleynets,li2018adaptive} and the spatial approaches~\cite{atwood2016diffusion,niepert2016learning,hamilton2017inductive,velivckovic2018graph,monti2017geometric,huang2018adaptive}.  

\subsubsection{Spectral Approaches}
Spectral approaches rely on the Laplacian spectrum to define graph convolution. 
For an undirected graph $\mathcal{G}=\{\mathcal{V}, \mathcal{E}\}$, 
$\mathbf{A}$ is the adjacency matrix, and $\mathbf{D}$ is the diagonal degree matrix, $\mathbf{D}_{ii}=\sum\nolimits_{j}^{N}\mathbf{A}_{ij}$. 
$\mathbf{L}= \mathbf{I}-\mathbf{D}^{-1/2}\mathbf{A}\mathbf{D}^{-1/2}$ represents the normalized Laplacian matrix of $\mathcal{G}$, and $\mathbf{L}$ can be decomposed as
$\mathbf{L}=\mathbf{U}\mathbf{\Lambda}\mathbf{U}^T$,
where $\mathbf{U}$ is the matrix of eigenvectors and $\mathbf{\Lambda}=\textit{diag}[\lambda_1,...,\lambda_N]$ is the diagonal matrix of eigenvalues.
Let $\mathbf{Z} \in \mathbb{R}^{N \times d}$ ($N=|\mathcal{V}|$) be the feature matrix of $\mathcal{G}$, and $\mathbf{z} \in \mathbb{R}^{N}$ be one of the columns of $\mathbf{Z}$ ($d=1$).
The graph Fourier transform of $\mathbf{z}$ is formulated as $\mathcal{F}(\mathbf{z})=\mathbf{U}^{T} \mathbf{z}$,  
and the inverse graph Fourier transform is $\mathcal{F}^{-1}(\hat{\mathbf{z}})=\mathbf{U} \hat{\mathbf{z}}$, where $\hat{\mathbf{z}} = \mathcal{F}(\mathbf{z})$.
Then, the convolution of $\mathbf{z}$ with a filter $\mathbf{g} \in \mathbb{R}^{N}$ is defined as 
$\mathbf{z} \ast_{\mathcal{G}} \mathbf{g}= \mathcal{F}^{-1}(\mathcal{F}(\mathbf{z})\odot\mathcal{F}(\mathbf{g}))=\mathbf{U}((\mathbf{U}^{T}\mathbf{z})\odot(\mathbf{U}^{T}\mathbf{g}))$, where $\ast$ is the graph convolution operator and $\odot$ is the Hadamard product.
By defining $\mathbf{g_\theta}=\textit{diag}(\mathbf{U}^{T}\mathbf{g})$, which is a function of $\mathbf{\Lambda}$, we have
\begin{equation}
	\mathbf{z} \ast_{\mathcal{G}} \mathbf{g}
 =\mathbf{U}\textit{diag}(\mathbf{U}^{T}\mathbf{g})\mathbf{U}^T\mathbf{z}
 =\mathbf{U}\mathbf{g_\theta} \mathbf{U}^T\mathbf{z}.
\end{equation}

\noindent 
\textbf{Chebyshev Spectral CNN (ChebNet)}~\cite{defferrard2016convolutional} uses Chebyshev polynomials to approximate the filtering operation $\mathbf{g}_\theta$.
$\mathbf{g}_\theta\approx\sum\nolimits_{i=0}^K\theta_iT_k(\mathbf{\tilde{L}})$, where $\tilde{\mathbf{L}}=2\mathbf{L}/\lambda_{max}-\mathbf{I}$ is the scaled Laplacian matrix, $\lambda_{max}$ is the largest eigenvalue of $\mathbf{L}$, and $\theta_i$'s are learnable parameters.
The Chebyshev polynomials can be defined recursively by $T_i(\mathbf{z}) = 2\mathbf{z}T_{i-1}(\mathbf{z})-T_{i-2}(\mathbf{z})$ with $T_0(\mathbf{z}) = 1$ and $T_1(\mathbf{z})=\mathbf{z}$. 
Then, the filtering operation is formulated as 
\begin{equation}
    \mathbf{z} \ast_{\mathcal{G}} \mathbf{g} 
    \approx \mathbf{U} \left( \sum\nolimits_{i=0}^{K} \theta_iT_i(\mathbf{\tilde{L}}) \right) \mathbf{U}^T\mathbf{z}
    \approx \sum\nolimits_{i=0}^{K} \theta_iT_i(\mathbf{\tilde{L}})\mathbf{z}.
\end{equation}


\noindent
\textbf{Graph Convolutional Networks (GCNs)}~\cite{kipf2017semi} 
introduce a first-order approximation of ChebNet ($K=1$). 
A GCN iteratively aggregates information from neighbors, and the feed forward propagation regarding node $v_i$ is conducted as
\begin{equation}
        \label{eq:GCN}
	\mathbf{h}^{(l+1)}_{i} = \sigma\left( \sum_{v_j \in \mathcal{N}(v_i)\cup\{v_i\}} \hat{a}(v_i, v_j) \mathbf{W}^{(l)}\mathbf{h}^{(l)}_{j} \right),
\end{equation}
where $\sigma(\cdot)$ is a nonlinear activation function, 
$\hat{\mathbf{A}}=(\hat{a}(v_i, u_j))$ denotes the re-normalized adjacency matrix $\mathbf{A}$, 
and $\mathbf{W}^{(l)}$ is a learnable transformation matrix in the $l$-th layer.
GCNs can also be interpreted from a spatial view~\cite{rong2020deep}.

\subsubsection{Spatial Approaches}
\textbf{GraphSAGE~\cite{hamilton2017inductive}} is a general inductive framework that updates node states by sampling and aggregating hidden states from a fixed number of local neighbors. 
Formally, it performs graph convolutions in the spatial domain, 
\begin{equation}
	\begin{split}
		\mathbf{h}^{(l+1)}_{\mathcal{N}_s(v_i)} &= \mathtt{Aggregator}_{l+1}\left( \{\mathbf{h}_{j}^{l}, \forall v_j \in \mathcal{N}_s(v_i)\} \right),  \\
		\mathbf{h}_{i}^{(l+1)} &= \sigma\left( \mathbf{W}^{(l+1)}\cdot[\mathbf{h}_{v_i}^{l}\oplus\mathbf{h}^{(l+1)}_{\mathcal{N}_s(v_i)}] \right),
	\end{split}
\end{equation}
where $\mathcal{N}_s(v_i)$ is a subset of nodes sampled from the full neighborhood $\mathcal{N}(v_i)$, and $\oplus$ is the concatenation operator.
As suggested by~\cite{hamilton2017inductive}, the aggregation function $\mathtt{Aggregator}_{l}$ can be the mean aggregator, max aggregator, LSTM aggregator, or pooling aggregator.

\noindent
\textbf{Graph Attention Networks (GAT)}~\cite{velivckovic2018graph}
introduces a self-attention mechanism to learn dynamic weights between connected nodes.   
It updates the hidden state of a node by attending to its neighbors,
\begin{equation}
	\footnotesize
	\begin{split}
        \mathbf{h}_{i}^{(l+1)} &= \sigma\left( \sum_{{v_j} \in \mathcal{N}(v_i)\cup\{v_i\}}\alpha_{ij}\mathbf{W}^{(l)}\mathbf{h}_{j}^{(l)} \right), \\
        \alpha_{ij} &= \frac{\exp\left( \mathtt{LReLU}(\mathbf{a}_{(l)}^T[\mathbf{W}^{(l)}\mathbf{h}_{i}^{(l)} \oplus \mathbf{W}^{(l)}\mathbf{h}_{j}^{(l)}] \right)}{\sum_{v_k \in \mathcal{N}(v_i)}\exp\left( \mathtt{LReLU}(\mathbf{a}_{(l)}^T[\mathbf{W}^{(l)}\mathbf{h}_{i}^{(l)} \oplus \mathbf{W}^{(l)}\mathbf{h}_{k}^{(l)}] \right)}
	\end{split}
\end{equation}
where 
$\alpha_{ij}$ is the pairwise attention weight and $\mathbf{a}$ is a vector of learnable parameters.
To increase the model's capacity and make the process of self-attention stable, GAT employs multi-head self-attention in practice. 

\subsubsection{New GNN Techniques}
\textbf{Deeper GNNs.}
A few recent works~\cite{xu2018representation,li2019deepgcns,rong2020dropedge} delve into building deep GCNs for a variety of basic graph-oriented tasks, such as node prediction and link prediction. 
DeepGCNs~\cite{li2019deepgcns} introduce commonly used concepts in CNNs, \emph{i.e.,} residual connections, dense connections, and dilated convolutions, to make GCNs go deeper as CNNs, such as a 56-layer GCN for point cloud semantic segmentation.
To alleviate the over-fitting and over-smoothing problems of deep GCNs, DropEdge~\cite{rong2020dropedge} proposes to randomly drop out a certain proportion of edges of the input graph for each training iteration.    
Moreover, Li~\emph{et al.}~\cite{li2021training} systemically investigate the effects of reversible connections, group convolutions, weight tying, and equilibrium models for improving the memory and parameter efficiency of GNNs, empirically revealing that combining reversible connections with deep network architectures could enable the training of extremely deep and wide GNNs, such as 1001 layers with 80 channels each and 448 layers with 224 channels each. 

\noindent
\textbf{Graph Pooling.} 
Graph pooling is a critical operation for modern GNN architectures. 
Inspired by the traditional CNN-based pooling, 
existing methods typically formulate graph pooling as a cluster assignment problem and explore the concept of local patches in the context of graphical structures.
Defferrard~\emph{et al.}~\cite{defferrard2016convolutional} achieve pooling with pre-defined subgraphs produced by a graph cut algorithm.
DiffPool~\cite{ying2018hierarchical} is a differentiable graph pooling module capable of generating hierarchical graph representations, seamlessly integrating with various GNN architectures in an end-to-end manner.
EigenPooling~\cite{ma2019graph} incorporates node features and local structures to obtain better assignment matrices.
Graph U-Nets~\etal~\cite{gao2019graph} presents a U-shaped architecture to implement pooling and up-sampling operations for GNNs.

\noindent
\textbf{Vision GNN.} ViG~\cite{han2022vision} directly represents an image as a graph, 
aiming to learn graph-level features for downstream vision tasks. They first split the input image into a set of regularly-shaped patches and regards each patch as a graph node. 
Graph edges are constructed using $K$-nearest neighbors of each node.
Then, multi-head graph convolution and positional encoding are performed at every node to jointly learn the topological structure and node features, and a feed-forward network (FFN) is used to mitigate the over-smoothing of node features and enhance the feature transformation capacity. 
In experiments, ViG outperforms DeiT by 1.7\% (top-1 accuracy) on ImageNet classification and Swin-T by 0.3\% (mAP) on MSCOCO object detection. 

\vspace{-3mm}

\subsection{Graph Transformers for 3D Data}
\textbf{Point Transformer}~\cite{zhao2021point} designs a local vector self-attention mechanism for point cloud analysis. In contrast, a related work, Point Cloud Transformer~\cite{guo2021pct}, adopts global attention. The local vector self-attention operation in each Transformer block of the Point Transformer is defined as, 
\begin{align}\nonumber
    \scriptsize 
    \begin{split}
    &\mathbf{h}^{(l+1)}_i = \\
    &\sum_{v_j \in \mathcal{N}_s(v_i)}\rho\left( \gamma\left( \mathbf{W}_Q^{(l+1)} \mathbf{h}_i^{(l)}-\mathbf{W}_K^{(l+1)} \mathbf{h}_j^{(l)}+\delta \right) \right) \odot \left( \mathbf{W}_V^{(l+1)} \mathbf{h}_j^{(l)}+\delta \right),
    \end{split}
\end{align}
where $\mathbf{W}_Q, \mathbf{W}_K, \mathbf{W}_V$ are shared parameter matrices to compute the query, key and value for attention-based aggregation, $\odot$ is the element-wise product, 
$\delta$ is a position encoding function, 
$\gamma$ is a nonlinear mapping function (\emph{e.g.} MLP),
and $\rho$ is a normalization function (\emph{e.g.} softmax). 
Recently, Fast Point Transformer~\cite{park2022fast} introduces a lightweight local self-attention architecture with voxel hashing to significantly improve efficiency. 
Stratified Transformer~\cite{lai2022stratified} samples distant points as additional keys to enlarge the receptive field, thereby modeling long-range dependencies.
Point Transformer V2~\cite{wu2022point} introduces grouped vector attention, improved position encoding, and partition-based pooling to enhance efficiency.

\noindent
\textbf{Mesh Graphormer}~\cite{lin2021mesh} develops a graph Transformer for mesh reconstruction from images. 
It exploits graph convolution and self-attention to learn local interactions within neighborhoods and non-local relations, respectively. 
Each Graphormer encoder block consists of five components, 
\emph{i.e.,} a Layer Norm, a Multi-Head Self-Attention~(MHSA) module, a Graph Residual Block, a second Layer Norm, and an MLP at the end.
Specifically, the MHSA module with $P$ heads accepts an input sequence $\mathbf{H}=\{\mathbf{h}_i\} \in \mathbb{R}^{n \times d}, i \in \{1,2,\cdots,n\}$ of $n$ tokens, and outputs $\{\mathbf{h}_i^p\}$ for each token, where $p \in \{1,2,\cdots,P\}$ is the head index.
Each $\mathbf{h}_i^p$ is computed as 
\begin{equation}
    \mathbf{h}_i^p = \mathbf{Att}\left( \mathbf{Q}_i^p, ~\mathbf{K}^p \right) \cdot \mathbf{V}^p \quad \in \mathbb{R}^{\frac{d}{P}},
\end{equation}
where $\mathbf{Q}$, $\mathbf{K}, \mathbf{V}$ are computed as $\mathbf{H}\mathbf{W}_Q$, $\mathbf{H}\mathbf{W}_K$, $\mathbf{H}\mathbf{W}_V$, respectively, $\mathbf{W}_Q$, $\mathbf{W}_K$, $\mathbf{W}_V$ are trainable parameter matrices in each layer, $\mathbf{Att}(\cdot)$ computes the original self-attention in \cite{vaswani2017attention}, and $\mathbf{Q}_i^p \in \mathbb{R}^{\frac{d}{P}}$, $\mathbf{K}^p, \mathbf{V}^p \in \mathbb{R}^{n \times \frac{d}{P}}$. 
Graph convolution (\emph{i.e.} Eq.~\ref{eq:GCN}) with a residual connection is further applied to $\mathbf{H}$.

\noindent
\textbf{GNNs \emph{v.s.} Vision Transformers (ViTs).} 
Given that ViTs resemble GNNs, especially GAT, in accounting for relations among spatially distributed entities, ViTs can be perceived as a special case of GNNs. 
Technically, however,
GNNs 
can use arbitrary relational inductive biases via a graph, \emph{i.e.,} we possess the flexibility to design any form of connectivity (regular or irregular connectivity) and are capable of incorporating multiple types of relations concurrently.
In comparison, ViTs focus on modeling relations 
in fully-connected global or local graphs. 
Therefore, if the data inherently exhibits a graph structure with irregular connectivity, such as scene graphs, GNNs are often a good choice.

\section{2D Natural Images}
\label{sec:2D}

\begin{figure}[h]
	\centering
	\includegraphics[width=0.4\textwidth]{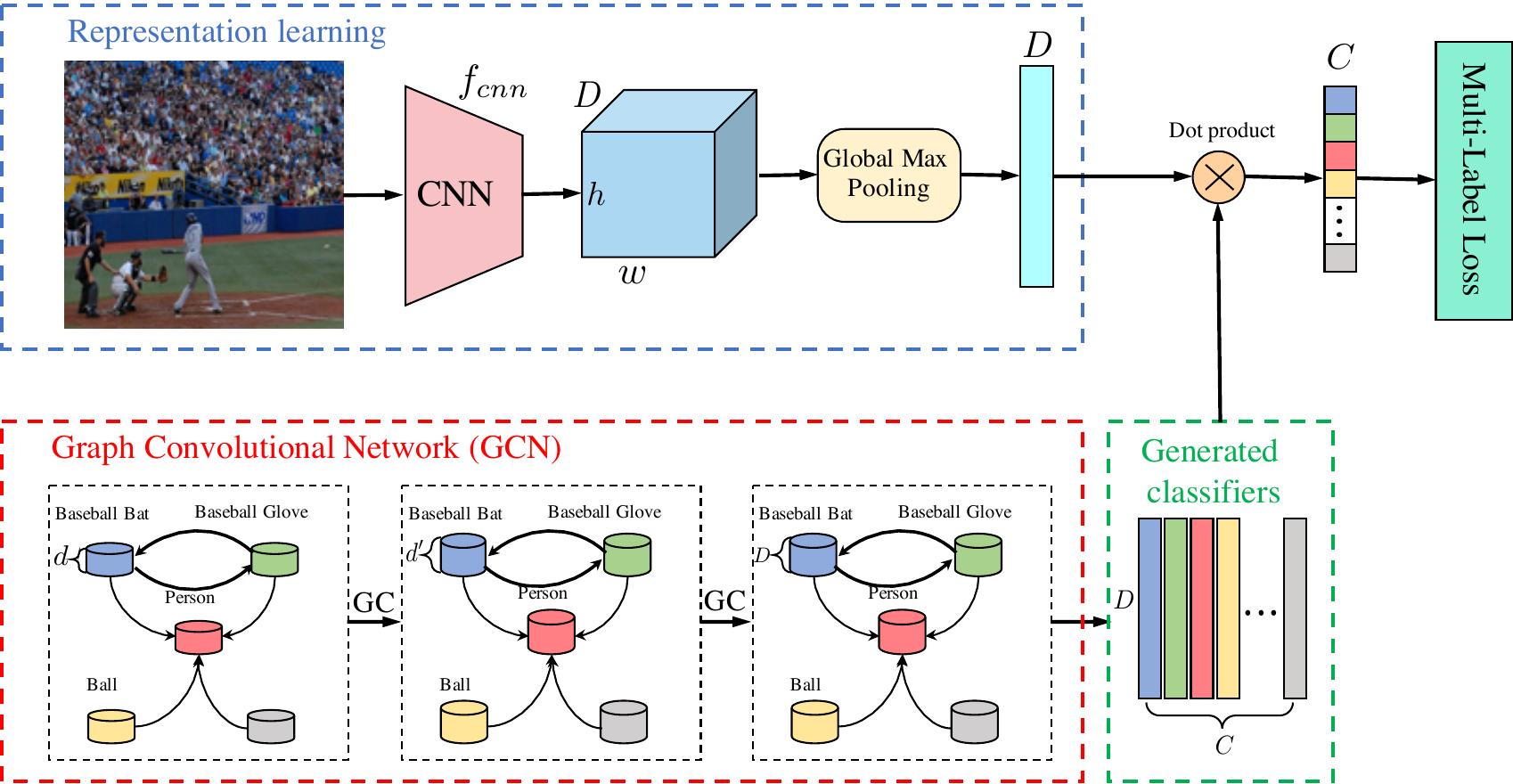}
	\vspace{-0.2cm}
	\caption{ML-GCN (Figure used courtesy of~\cite{chen2019multi}).}\label{fig:ML-GCN}
    \vspace{-0.5cm}
\end{figure}

\subsection{Image Classification}
Thanks to the strong relational inductive biases provided by GNNs, \emph{e.g.,} arbitrary connectivity patterns (one-to-many and many-to-many), recent approaches focus on modeling structural dependencies among different categories. 
This enables better visual and semantic understanding and improves transparency of the prediction process and the overall performance. Given a single or a set of image(s) $I$, we have
\begin{align}\small
    \mathcal{G}(\mathcal{V}, \mathcal{E}, \mathbf{X}) = \Pi(I)
\end{align}
where $\Pi$ represents the graph construction process, and $\mathbf{X}$ is the feature matrix. 
Then, a common choice to define the adjacency matrix $\mathbf{A}$ based on $\mathcal{G}$ is Radial Basis Function, 
\begin{align}\small
    \mathbf{A}_{ij} = \exp{\left(-\frac{d(\mathbf{x}_i, \mathbf{x}_j)}{2\delta^2}\right)}
\end{align}
where $d(\cdot,\cdot)$ is a distance measure (\emph{e.g.,} Euclidean distance and cosine similarity), $\delta$ is a scaling factor, and $\mathbf{x}$ can be either a class prototype or a specific instance.
Note that different types of adjacency matrices, \emph{e.g.,} language-based label dependencies and image structural priors, can be jointly considered. In addition to using pre-defined adjacency matrices, we can also opt for learnable ones by introducing additional parameters. 
These matrices will be merged into a unified matrix.
After that, we can perform learning process, 
\begin{align}\small
   \hat{\mathbf{X}} = \text{GNN}(\mathcal{G}, \mathbf{A})
\end{align}
where $\hat{\mathbf{X}}$ is the updated feature matrix which can be projected to the visual domain, combined with other feature extraction modules, or directly used for task-specific prediction.

\subsubsection{Multi-Label Classification}
This task aims to recognize a set of objects within a single image. 
The key idea for GNN-based methods is to construct a per-image label graph for relational modeling and reasoning, ensuring that different categories are no longer isolated. 
Semantic object representations and label relationships can be learned simultaneously or independently.
For the former, a representative work~\cite{chen2019multi} (cf. Fig.~\ref{fig:ML-GCN}) builds a directed graph over the label space, where each node is a word (label) embedding, and the connections denote their relations.
GCN maps the constructed graph into a set of interdependent binary classifiers, 
and a label correlation matrix is designed to guide information propagation among nodes.
For the latter,
Chen~\emph{et al.}~\cite{chen2019learning} introduced statistical label co-occurrence to directly guide the propagation of semantic features of different object regions.
Follow-up approaches~\cite{lanchantin2019neural,wu2020adahgnn,ye2020attention,you2020cross,nguyen2021modular,zhao2021transformer} focus on modeling semantic label dependencies via more elaborate architectures, such as hypergraph neural networks~\cite{wu2020adahgnn}, cross-modality attention~\cite{you2020cross}, graph Transformer~\cite{nguyen2021modular}, and Transformer~\cite{zhao2021transformer}.
The performance of these algorithms is summarized in Tab.~\ref{tab:multi-label}.

\begin{center}
	\begin{table}[t]
		\caption{Performance (mAP) on two multi-label classification benchmarks, \emph{i.e.,} MS-COCO and VOC 2007.}
		\label{tab:multi-label}
		\vspace{-0.2cm}
		\centering
		\setlength\tabcolsep{5pt}
		\begin{tabular}{cc|cc}
			\toprule
			Method & Reference &  MS-COCO & VOC 2007 \\
			\midrule
			ResNet-101~\cite{he2016deep} & CVPR'16 & 77.3 & 89.9 \\
			ML-GCN~\cite{chen2019multi} & CVPR'19 & 83.0 & 94.0 \\
			SSGRL~\cite{chen2019learning} & ICCV'19 & 83.8 & 93.4 \\
			MS-CMA~\cite{you2020cross} & AAAI'20 & 83.8 & - \\
			ADD-GCN~\cite{ye2020attention} & ECCV'20 & 85.2 & 93.6 \\
			TDRG~\cite{zhao2021transformer} & ICCV'21 & 86.0 & 95.0 \\
			\bottomrule
			\vspace{-8mm}
		\end{tabular}
	\end{table}
\end{center}

\vspace{-1cm}
\subsubsection{Few-Shot Learning~(FSL)}
FSL aims to generalize to new tasks that only have a few samples. 
GNN was introduced to compensate for the lack of semantic correlations.
TPN~\cite{liu2019learning} constructs graphs in the embedding space to exploit the manifold structure of novel classes. 
Label information is propagated from the support set to the query set based on the constructed graphs.
Bottom-up and top-down reasoning modules~\cite{chen2021hierarchical} are introduced to explore the hierarchical relations of semantic classes.  
Compared to the node-based labeling frameworks, 
\cite{kim2019edge} proposes an edge-labeling GNN that learns to predict edge labels, explicitly constraining intra- and inter-class similarities.
Instead of using a GNN to perform label propagation,
Yu~\etal~\cite{yu2022hybrid} introduce an instance GNN and a prototype GNN as feature embedding task adaptation modules for quickly adapting learned features to new tasks. 
Moreover, meta-learning-based approaches~\cite{garcia2018few,yang2020dpgn,luo2020learning} can directly solve data scarcity by imitating the distribution shift during training and performing class-based information transfer.

\subsubsection{Zero-Shot Learning (ZSL)}
ZSL aims to classify samples from classes that have not been seen during training, and predefined or learnable knowledge graphs are used to represent semantic relations. 
Due to the absence of training data, 
they resort to certain prior knowledge, such as language-based relation graphs, to assist in understanding information unseen during training.
To enhance knowledge propagation across distant nodes in a graph network, \cite{kampffmeyer2019rethinking} leverages the hierarchical structure of the graph, such as the relations between different levels of species classification.
To promote structural knowledge transfer (\emph{e.g.} semantic descriptions of classes) from seen to unseen classes,
Xie~\etal~\cite{xie2020region} introduce a region-based graph to model the visual relations among different regions within an input image, which are expected to generalize well on unseen classes. 
Instead of using pre-defined attributes to bridge seen and unseen classes, 
\cite{liu2020attribute} resort to the inter-class relations to generate attribute vectors and propagate the dependencies among them via graph convolution.
Recent studies strive to improve the efficacy of knowledge propagation across different classes, such as joint visual and semantic prototype propagation on auto-generated graphs~\cite{liu2021isometric}, GATs for exploiting appearance relations among local regions~\cite{chen2022gndan}, as well as explicit compositional relation modeling~\cite{mancini2022learning}.   

\subsubsection{Transfer Learning}
Domain Adaptation (DA) and Domain Generalization (DG), both sub-branches of transfer learning, benefit from knowledge graphs depicting relations among classes.
Such structural knowledge is expected to be transferable across domains and thus can be reused for novel environments.
Wang~\etal~\cite{ma2019gcan} propose an adversarial GCN for DA, where a GCN is established over densely connected instance graphs using mini-batch samples to encode data structure information. 
For DG, Chen~\etal~\cite{chen2022compound} build global prototypical relation graphs and introduce a graph self-attention mechanism to model long-range dependencies among different categories. 
Other works introduce more elaborate training algorithms to aggregate and propagate information in both intra- and inter- domains, such as curriculum learning to achieve progressive aggregation~\cite{roy2021curriculum}, progressive graph learning to select reliable pseudo-labels~\cite{luo2020progressive}, prototype alignment to assist the structural knowledge transfer between domains~\cite{wang2020prototype,wang2020learning}.

\noindent
\textbf{Discussion.}
GNNs used for image classification have been mostly explored for multi-object, data- or label-efficient scenarios (\emph{e.g.,} zero-shot and few-shot learning), aiming to model the complex relations among different objects/classes for compensating the shortage of training samples or supervision signals.
Current work focuses on extracting ad-hoc knowledge graphs from the data for a certain task,
which is heuristic and relies on the human prior. Future work is expected to \emph{(i)} develop general and automatic graph construction procedures, \emph{(ii)} enhance the interactions between abstract graph structures and task-specific classifiers, and \emph{(iii)} excavate more fine-grained building blocks (node and edge) to increase the capability of constructed graphs.

\subsection{Object Detection}
Object detection~\cite{liu2020deep} 
aims to localize and recognize all object instances of given classes in input images.
Despite the fruitful progress, 
modern object detectors often overlook the relationships and interactions among object instances by treating each object individually.
Thus, two challenges arise: 
(1) reasoning over semantic dependencies, co-occurrence, and relative locations of objects in addition to perceiving individual objects, and
(2) embedding object-level dependencies into the detection pipeline when data distributions exhibit structural object relations. In general, given an input image $I$ with a set of objects $\mathcal{O}$, the goal of GNN-based methods is to perform both local and global relational reasoning over these objects or parts of objects,
\begin{align}
    \mathcal{G}=\Pi(I, \mathcal{O}),\;\hat{\mathbf{X}} = \text{GNN}(\mathcal{G}, \mathbf{A}).
\end{align}

\begin{figure}[!t]
	\centering
	\includegraphics[width=0.48\textwidth]{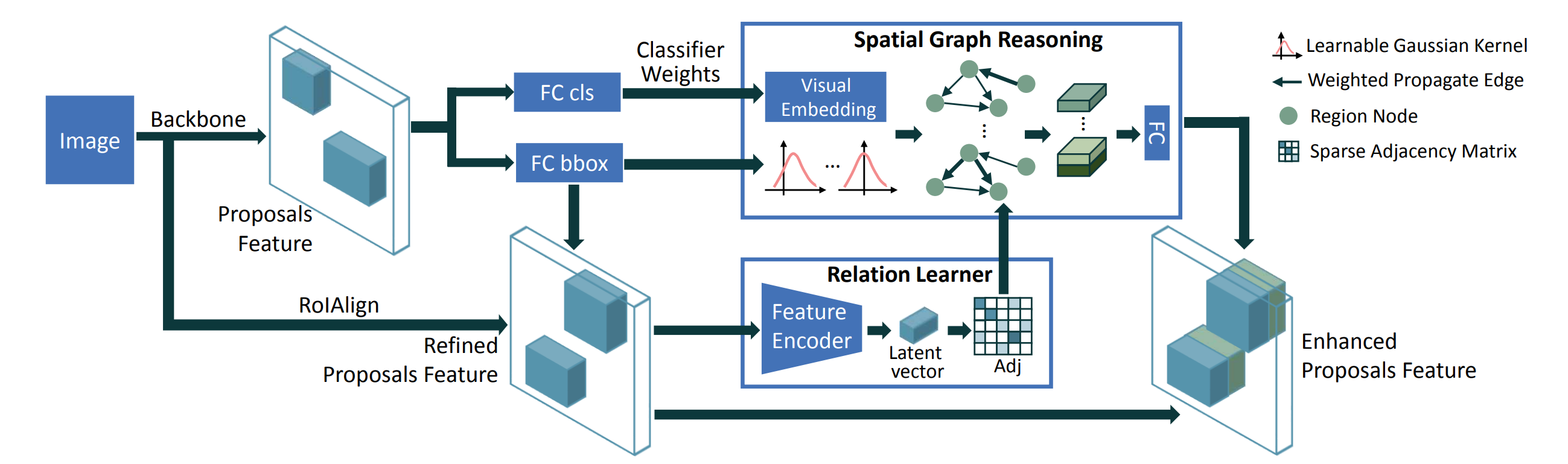}
	\vspace{-0.2cm}
	\caption{SGRN (Figure used courtesy of~\cite{xu2019spatial}).}\label{fig:SGRN}
	\vspace{-0.5cm}
\end{figure}

Reasoning-RCNN~\cite{xu2019reasoning} presents an adaptive global reasoning network for large-scale object detection by incorporating commonsense knowledge (category-wise knowledge graph) and propagating visual information globally, thereby endowing detection networks with the capability of reasoning.
SGRN~\cite{xu2019spatial} (cf. Fig.~\ref{fig:SGRN}) goes one step further by adaptively discovering semantic and spatial relationships without requiring prior handcrafted linguistic knowledge.
The relation learner learns a sparse adjacency matrix to encode contextual relations among different regions.
A spatial graph reasoning module utilizes a learned adjacency matrix and learnable spatial Gaussian kernels to perform feature aggregation and relational reasoning.
RelationNet~\cite{hu2018relation} proposes an adapted attention module for detection head networks, explicitly learning relations among objects through encoding long-range dependencies.
RelationNet++~\cite{chi2020relationnet++} presents a self-attention based decoder module to embrace the strengths of different object/part representations within a single detection framework.
Li~\emph{et al.}~\cite{li2020gar} introduce a heterogeneous graph to jointly model object-object and object-scene relations.
Recently, GraphFPN~\cite{zhao2021graphfpn} presents a graph feature pyramid network, which explores contextual and hierarchical structures of an input image using a superpixel hierarchy, enabling within-scale and cross-scale feature interactions with the help of spatial and channel attention mechanisms. 


In addition to improving the in-distribution performance,
domain adaptive object detection (DAOD)
has received a great deal of attention from many real-world applications.
A natural choice for modeling cross-domain relations is a bipartite graph $\mathcal{G} = \{\mathcal{V}_s, \mathcal{V}_t, \mathcal{E}\}$, where vertices are divided into two disjoint sets and edges connect vertices from different sets.
In practice, Chen~\etal~\cite{chen2022relation,chen2021dual} first builds intra- and inter-domain relation graphs in virtue of cyclic between-domain consistency without assuming any prior knowledge about the target distribution. It then incorporates bipartite GCNs and graph attention mechanisms to model homogeneous and heterogeneous object dependencies and interactions in both pixel and semantic spaces. 
In addition, SIGMA~\cite{li2022sigma} formulates DAOD as a graph matching problem by setting up cross-image graphs to model class-conditional distributions on both domains.
SRR-FSD~\cite{zhu2021semantic} introduces a semantic relation reasoning module, where each class has a corresponding node in a dynamic relation graph, to integrate semantic relations between base and novel classes for novel object detection. By doing so, semantic information is propagated through graph nodes, endowing an object detector with semantic reasoning ability.

\noindent
\textbf{Discussion.}
Owing to the long-range modeling capability,
GNN-based detection methods exploit between-object~\cite{xu2019spatial}, cross-scale~\cite{zhao2021graphfpn} or cross-domain~\cite{chen2022relation} relationships, as well as relationships between base and novel classes~\cite{zhu2021semantic}. Due to the introduction of additional learning modules, however, the optimization process becomes harder than vanilla detection models. Moreover, the compatibility between Euclidean and non-Euclidean structures will be challenged if not properly harmonized.
In future, researchers could \emph{(i)} design better region-to-node feature mapping methods, \emph{(ii)} incorporate Transformer (or pure GNN) encoders to improve the expressive power of initial node features, and \emph{(iii)} instead of resorting to forward and backward feature mappings, directly perform reasoning in the original feature space to better preserve the intrinsic structure of images.
%


\subsection{Image Segmentation}
Semantic segmentation divides an image into several semantically meaningful regions to perform pixel-wise labeling. 
While CNN-based architectures have made significant advancements, their intrinsic limitations, such as the inability to reason about distant regions of arbitrary shapes, pose challenges in achieving a holistic understanding of a scene. In this regard, GNNs offer a unified framework for modeling both object appearances $\mathcal{R}_a$ and image contexts $\mathcal{R}_c$.
\begin{align}
    \mathcal{G}=\Pi(I),\;\mathbf{A} = \mathtt{E}(\mathcal{R}_a, \mathcal{R}_c),\;\hat{I} = \mathtt{P}(\text{GNN}(\mathcal{G}, \mathbf{A})),
\end{align}
where $\mathtt{E}(\cdot)$ is a function that encodes both $\mathcal{R}_a$ and $\mathcal{R}_c$, $\mathtt{P}(\cdot)$ is a pixel-wise predictor, and $\hat{I}$ represents prediction results.



A common goal is to globally model contextual and semantic relations in the backbone feature space.
Zhang~\etal~\cite{zhang2019dual} propose a dual GCN framework, where a coordinate space GCN models spatial relations among pixels, and a feature space GCN models dependencies among channel dimensions of a feature map.
After that, these two features are mapped back to the original coordinate space.
To improve local feature aggregation,
Chen~\etal~\cite{chen2019graph} design a global reasoning unit by projecting features that are globally aggregated in the coordinate space to the node domain and performing relational reasoning in a fully connected graph.
To model long-range dependencies and avoid constructing fully connected graphs,
DGMN~\cite{zhang2020dynamic} dynamically samples the neighborhood of a node and predicts node dependencies, filter weights, and affinities for information propagation.
Similarly, Yu~\etal~\cite{yu2020representative} dynamically sample representative nodes for relational modeling.
Instead of constructing additional semantic interaction spaces (projection and re-projection),
Li~\etal~\cite{li2020spatial} propose an improved Laplacian formulation that enables graph reasoning in the original feature space, fully exploiting contextual relations at different feature scales.
Hu~\etal~\cite{hu2020class} introduce a class-wise dynamic graph convolution module to perform graph reasoning over pixels that belong to the same class to dynamically aggregate features.
We summarize the performance in Tab.~\ref{tab:segmentation}.

For one-shot semantic segmentation,
Zhang~\emph{et al.}~\cite{zhang2019pyramid} introduce a pyramid graph attention module to model the connection between query and support feature maps, associating unlabeled pixels in the query set with semantic and contextual information.
For few-shot semantic segmentation,
Xie~\emph{et al.}~\cite{xie2021scale} propose a scale-aware GNN to perform cross-scale relational reasoning over both support and query images. A self-node collaboration mechanism is introduced to perceive different resolutions of the same object.
Zhang~\emph{et al.}~\cite{zhang2021affinity} propose an affinity attention GNN for weakly supervised semantic segmentation.
Specifically, an image is first converted to a weighted graph via an affinity CNN network, and then an affinity attention layer is devised to model long-range interactions in the constructed graph and propagate semantic information to unlabeled pixels.
In addition to semantic segmentation,
Wu~\emph{et al.}~\cite{wu2020bidirectional} design a bidirectional graph reasoning network to bridge a pair of ``things'' and ``stuff'' branches for panoptic segmentation.
They first build things and stuff graphs to represent relations within corresponding branches. Bidirectional graph reasoning is then performed to propagate semantic representations both within and across the two branches.

\begin{center}
	\begin{table}[!t]
		\caption{\small Performance (mIOU) on segmentation benchmarks, \emph{i.e.,} Cityscapes, PASCAL-Context, and COCO Stuff.}
		\label{tab:segmentation}
		\vspace{-0.2cm}
		\centering
		\setlength\tabcolsep{3pt}
            \scalebox{0.85}{
		\begin{tabular}{cc|ccc}
			\toprule
			Method & Reference &  Cityscapes & PASCAL & COCO Stuff \\
			\midrule
			DGCNet~\cite{zhang2019dual} & BMVC'19 & 80.5 & 53.7 & - \\
			GloRe~\cite{chen2019graph} & CVPR'19 & 80.9 & - & - \\
			DGMN~\cite{zhang2020dynamic} & CVPR'20 & 81.6 & - & - \\
			SpyGR~\cite{li2020spatial} & CVPR'20 & 81.6 & 52.8 & 39.9 \\
                RGNet~\cite{yu2020representative} & ECCV'20 & 81.5 & 53.9 & - \\
                CDGCNet~\cite{hu2020class} & ECCV'20 & 82.0 & - & 40.7 \\
			\bottomrule
		\end{tabular}}
            \vspace{-6mm}
	\end{table}
\end{center}

\vspace{-0.8cm}

\noindent
\textbf{Discussion.}
Many research efforts in semantic segmentation are devoted to exploring contextual information in the local- or global-level with pyramid pooling, dilated convolutions, or the self-attention mechanism.
For instance, non-local networks~\cite{wang2018non} and their variants achieve this goal by adopting the self-attention mechanism, but are computationally expensive as comparing every pixel to all other pixels in an image has a quadratic time complexity.
In addition, probabilistic graphical models, such as Conditional Random Fields (CRFs) and Markov Random Fields (MRFs), can be used to model scene-level semantic contexts but have very limited representation ability.
In comparison to these prior efforts, GNN-based methods show clear superiority in terms of relation modeling and training efficiency.

\begin{figure}[!t]
	\centering
	\includegraphics[width=0.48\textwidth]{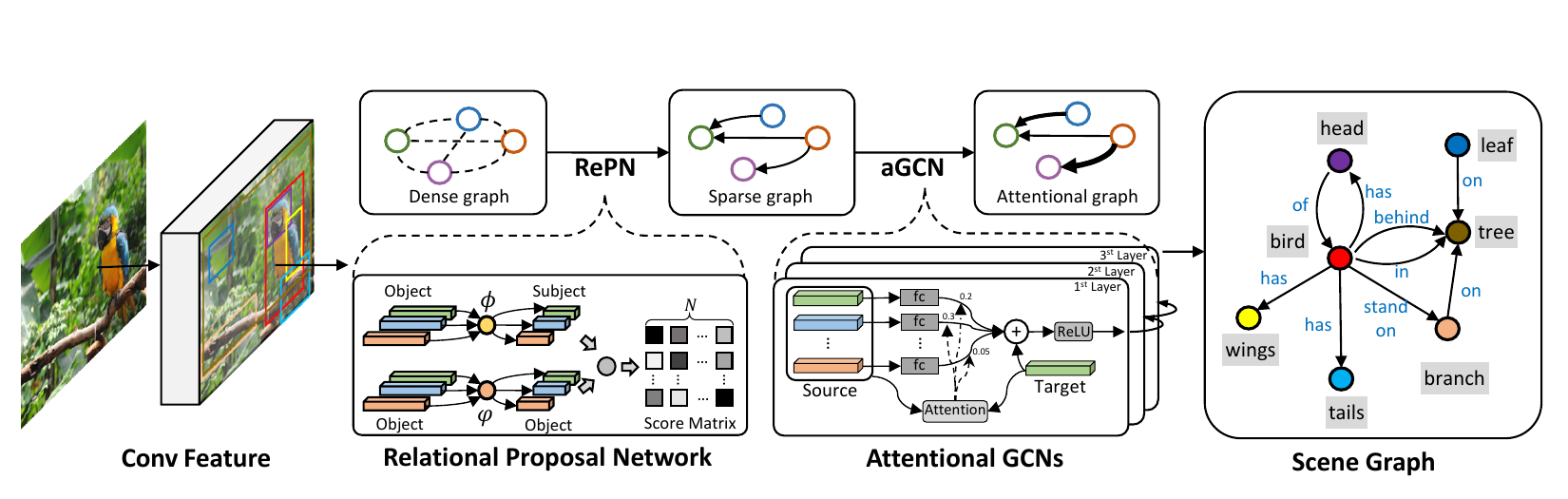}
	\caption{Graph R-CNN (Figure used courtesy of~\cite{yang2018graph}).}\label{fig:yang2018graph}
	\vspace{-5mm}
\end{figure}

\vspace{-2mm}

\subsection{Scene Graph Generation (SGG)}
SGG~\cite{johnson2015image,zhu2022scene} refers to the task of detecting object pairs and their relations in an image to generate a visually-grounded scene graph, which provides a high-level understanding of a visual scene.
A scene graph is the result of image parsing by representing objects in the image as nodes and their relations as edges. 
This aligns well with the nature of GNNs, which do not require prior knowledge to construct task-specific graph structures from inputs, unlike other computer vision tasks.
GNN-based methods for SGG typically consist of three stages: object detection, relation graph construction and refinement, and relation prediction, \emph{i.e.,}
\begin{align}
    P(\mathcal{S}|I) = P(\mathcal{V}|\mathcal{I})P(\mathcal{E}|\mathcal{V},I)P(\mathcal{R}, \mathcal{O}|\mathcal{V}, \mathcal{E}, I),
\end{align}
where $I$ is an image, $\mathcal{S}=(\mathcal{V}, \mathcal{E}, \mathcal{O}, \mathcal{R})$ is a scene graph, and $\mathcal{O}$ and $\mathcal{R}$ are object and relationship labels respectively.

Their key idea is to align visual and textual entities (including their topological homogeneity) in a shared latent space.
Graph R-CNN~\cite{yang2018graph} (cf. Fig.~\ref{fig:yang2018graph}) first obtains a sparse candidate graph by pruning the densely connected graph generated from RPN via a relation proposal network, then an attentional GCN is introduced to aggregate contextual information and update node features and edge relations.
Li~\emph{et al.}~\cite{li2018factorizable} 
utilizes a spatially weighted message-passing structure to refine features of objects and subgroups by passing messages among them with attention-like schemes.
Qi~\emph{et al.}~\cite{qi2019attentive} propose attentive relational networks, which first transform label embeddings and visual features into a shared semantic space, and then rely on a GAT to perform feature aggregation for final relation inference.
Li~\etal~\cite{li2021bipartite} use a bipartite GNN to estimate and propagate relation confidence in a multi-stage manner.
Suhail~\etal~\cite{suhail2021energy} propose an energy-based framework, which relies on a graph message passing algorithm to compute the energy of configurations.

\noindent
\textbf{Discussion.} Due to the inherent connection between SGG and GNNs, many advanced graph representation paradigms can be effectively applied to SGG.
However, current approaches in SGG often adopt a stage-wise manner, and exploring methods that can directly generate a scene graph in a single stage is an open question. Further research can focus on developing novel generative models, such as graph-based diffusion models, to better exploit the relational inductive bias in generating scene graphs.

\section{Video Understanding}
\label{sec:video}

Researchers have explored utilizing GNNs in video understanding tasks, including video action recognition~\cite{wang2018videos, yan2018spatial}, temporal action localization~\cite{zeng2019graph, xu2020g}, and video object segmentation~\cite{wang2019zero, lu2020video}. 
As spatio-temporal relations are critical in video understanding, GNNs are naturally applied to perform spatial and semantic relation reasoning over time among visual constituents. 
Specifically, GNNs are adapted in video understanding tasks to facilitate modeling (i) spatio-temporal relations of objects among single or multiple frames, ranging from first-order relations to higher-order relations~\cite{wang2018videos}, (ii) dependencies between frames at multiple time scales~\cite{zhang2020temporal}, (iii) dependencies among joints in the same or consecutive frames in skeleton-based approaches~\cite{yan2018spatial}. 

\subsection{Video Action Recognition}
Video human action recognition is one of the fundamental tasks in video processing and understanding, which aims to identify and classify human actions in RGB/depth videos or skeleton data.
Regardless of the data modality, modeling spatio-temporal contexts using humans, objects and joints is critical in identifying human action.
Typically, one or more graphs $\mathcal{G}$ are constructed from objects, humans, video frames or skeletons. Then, GNNs are employed to model the relationships among these elements and make predictions regarding human actions as follows:
\begin{align}
    \small
        \mathcal{G} &= \Pi (\hat{\mathcal{V}}, \hat{\mathcal{E}}),\;
        \mathcal{F} = \text{GNN}(\mathcal{G}),
\end{align}
where 
$\hat{\mathcal{V}}$ denotes the nodes that can represent objects, humans, video frames, or skeletons, $\hat{\mathcal{E}}$ denotes the edges that can capture spatial relations between objects in a single frame or temporal relations among multiple frames, 
and $\mathcal{F}$ represents the aggregated information for final prediction. Some of these methods employ RNN-style structures for temporal modeling in this context.

\begin{figure}[!t]
	\centering
	\includegraphics[width=0.9\columnwidth]{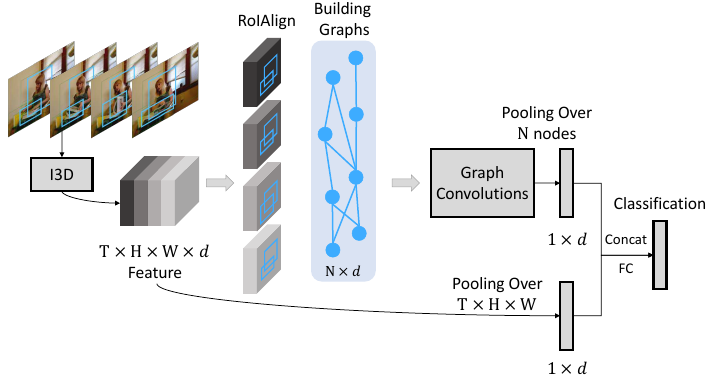}
	\caption{Long-range temporal modeling via GCNs (Figure used courtesy of~\cite{wang2018videos}).}
	\label{fig:video88}
    \vspace{-7mm}
\end{figure}

\noindent\textbf{Action Recognition.}
Wang {\it et al.}~\cite{wang2018videos} propose to capture long-range temporal contexts via graph-based reasoning over human-object and object-object relations. As shown in Fig.~\ref{fig:video88}, they connect all the humans and objects in all the frames to construct a space-time region graph, where edges based on appearance similarity connect every pair of objects while edges based on spatio-temporal relations connect spatially overlapping objects in consecutive frames. The results from the GCNs built on this graph are used to predict human action.
Ou {\it et al.}~\cite{ou2022object} improve the spatio-temporal graph in \cite{wang2018videos} by constructing actor-centric object-level graphs and applying GCNs to capture the contexts among objects in an actor-centric way. In addition, a relation-level graph is built to model the contexts of relation nodes. Instead of spatio-temporal modeling,
Zhang {\it et al.}~\cite{zhang2020temporal} propose multi-scale reasoning over the temporal graph of a video, where each node is a video frame, and the pairwise relations between nodes are represented as a learnable adjacency matrix. To capture both short-term and long-term dependencies, both GAT and the temporal adjacency matrix have multiple heads, each of which investigates one kind of temporal relations. 

Some other studies have explored alternative approaches, such as leveraging knowledge graphs for zero-shot action recognition~\cite{gao2019know} and employing graph-based higher-order relation modeling for long-term action recognition~\cite{zhou2021graph}. Besides human action recognition, GNNs have also been applied to group activity recognition~\cite{wu2019learning} and action performance assessment~\cite{pan2019action}.

\noindent\textbf{Skeleton-Based Action Recognition.} Skeletal data plays a vital role in action recognition because spatio-temporal dependencies among human body parts indicate human motion and action. Skeleton-based action recognition recognizes human actions in a skeleton sequence extracted from a video. Considering the human skeleton has a graph structure with natural joint connections,
Yan {\it et al.}~\cite{yan2018spatial} propose an ST-GCN network that first connects joints in the same frame according to the natural connectivity in the human skeleton, and then connects the same joints in two consecutive frames to maintain temporal information. They run GCNs on the joint graph to learn both spatial and temporal action patterns. As shown in Tab.~\ref{tab:video}, their ST-GCN network improves on previous methods significantly.
Shi {\it et al.}~\cite{shi2019two} improve~\cite{yan2018spatial} by introducing a fully-connected graph with learnable edge weights between joints and a data-dependent graph learned from the input skeleton. They build GCNs on all the three graphs.
Similar to \cite{shi2019two},
Li {\it et al.}~\cite{li2019spatio} and Li {\it et al.}~\cite{li2019actional} also connect physically-apart skeleton joints to capture the patterns of collaboratively moving joints.
Different from~\cite{li2019spatio},
Zhao {\it et al.}~\cite{zhao2019bayesian} improve joint connectivity in a single frame~\cite{yan2018spatial} by adding edges between limbs and head. In addition, they use GCNs to capture relations between joints in individual frames and adopt LSTM~\cite{hochreiter1997long} to capture temporal dynamics.
Instead of only updating node features in GCNs,
Shi {\it et al.}~\cite{shi2019skeleton} maintain edge features and learn both node and edge feature representations via directed graph convolution.

Unlike previous works that model local spatial and temporal contexts in consecutive frames,
Liu {\it et al.}~\cite{liu2020disentangling} capture long-range spatio-temporal dependencies. They first construct multiple dilated windows over the temporal dimension.
Then, in each window, separate GCNs are used on multiple graphs with different scales. Finally, the results of GCNs are aggregated on all the graphs in all the windows to capture multi-scale and long-range dependencies. Some additional studies~\cite{si2019attention, cheng2020skeleton, cheng2020decoupling} have explored combining or modifying standard graph convolution layers with other modules (\emph{e.g.,} LSTM and shift CNNs~\cite{wu2018shift}) to enhance their suitability for action recognition tasks.

\begin{table}
	\renewcommand\arraystretch{1.2}
	\centering
        \caption{Performance comparison on datasets for action recognition. GNN-based methods are marked with $^\star$.}
        \vspace{-3mm}
	\resizebox{\linewidth}{!}{%
		\begin{tabular}{l|c|c|c} 
			\hline
			Method                            & \multicolumn{1}{c|}{\begin{tabular}[c]{@{}c@{}}Charades~\cite{sigurdsson2016hollywood}\\mAP \end{tabular}} & \multicolumn{1}{c|}{\begin{tabular}[c]{@{}c@{}}NTU-RGB+D~\cite{shahroudy2016ntu}\\Accuracy(\%)\\X-Sub~~~ X-View\end{tabular}} & \begin{tabular}[c]{@{}c@{}}Kinetics Skeleton~\cite{kay2017kinetics}\\Accuracy(\%)\\Top1 ~ ~ Top5\end{tabular}  \\ 
			\hline
			\multicolumn{4}{l}{Action Recognition}                                                                                                                                                                                                                                                                                                                                                                                                            \\
			NL + I3D~\cite{wang2018non}          & 37.5                                                                        & -~ ~~ ~~~~~~~ -                                                                                         & -~~~~~~~~~~ -                                                                            \\
			STRG~\cite{wang2018videos}$^\star$    & 39.7                                                                        & -~ ~~ ~~~~~~~ -                                                                                         & -~~~~~~~~~~ -                                                                            \\
			$OR^2G$~\cite{ou2022object}$^\star$     & 44.9                                                                        & -~ ~~ ~~~~~~~ -                                                                                         & -~~~~~~~~~~ -                                                                            \\
			TS-GCN~\cite{gao2019know}$^\star$     & 40.2                                                                        & -~ ~~ ~~~~~~~ -                                                                                         & -~~~~~~~~~~ -                                                                            \\
			GHRM~\cite{zhou2021graph}$^\star$       & 38.3                                                                        & -~ ~~ ~~~~~~~ -                                                                                         & -~~~~~~~~~~ -                                                                            \\ 
			\hline
			\multicolumn{4}{l}{Skeleton-Based Action Recognition}                                                                                                                                                                                                                                                                                                                                                                                           \\
			Deep LSTM~\cite{shahroudy2016ntu}            & -                                                                           & 60.7~ ~ ~~~ 67.3                                                                                  & 16.4~~~~ 35.3                                                                      \\
			ST-GCN~\cite{yan2018spatial}$^\star$        & -                                                                           & 81.5 ~ ~~~~ 88.3                                                                                  & 30.7 ~~~ 52.8                                                                      \\
			2s-AGCN~\cite{shi2019two}$^\star$     & -                                                                           & 88.5 ~ ~~~~ 95.1                                                                                  & 36.1 ~~~ 58.7                                                                      \\
			STGR~\cite{li2019spatio}$^\star$           & -                                                                           & 86.9~ ~ ~ ~ 92.3                                                                                  & 33.6 ~~~~ 56.1                                                                     \\
			AS-GCN~\cite{li2019actional}$^\star$        & -                                                                           & 86.8 ~ ~~ ~ 94.2                                                                                  & 34.8~~~~~ 56.5                                                                     \\
			Zhao et al.~\cite{zhao2019bayesian}$^\star$         & -                                                                           & 81.8 ~ ~~ ~ 89.0                                                                                  &   -~~~~~~~~~~ -                                                                   \\
			DGNN~\cite{shi2019skeleton}$^\star$            & -                                                                           & 89.9~~ ~ ~~ 96.1                                                                                  & 36.9~~~~~ 59.6                                                                     \\
			MS-G3D Net~\cite{liu2020disentangling}$^\star$        & -                                                                           & 91.5 ~ ~ ~~ 96.2                                                                                  & 38.0 ~~~~ 60.9                                                                     \\
			\hline
		\end{tabular}
	}
	\label{tab:video}
	\vspace{-0.5cm}
\end{table}


\subsection{Temporal Action Localization}
Temporal action localization aims to localize temporal intervals and recognize action instances in an untrimmed video. Video contexts play an important role in action detection as they can be used to infer potential actions~\cite{xu2020g}. This section reviews methods that utilize GNNs to obtain video contexts.
Zeng {\it et al.}~\cite{zeng2019graph} follow a two-stage pipeline for temporal action localization, which first generates temporal proposals, then classifies them and regresses their temporal boundaries. In the second stage, pairs of proposals are connected according to their temporal intersection over union and the $L_1$ distance over union to form a proposal graph. GCNs are run on the graph to capture the relations among proposals, and the output from the GCNs is used to predict the boundary and category of every proposal again.
Unlike~\cite{zeng2019graph} taking temporal proposals as nodes, Xu {\it et al.}~\cite{xu2020g} represent a video sequence as a snippet graph, where each node is a snippet and each edge corresponds to the correlation between two snippets. Both temporal edges between snippets in two consecutive time steps and semantic edges learned from snippet features are used to build graph connectivity. A dynamic graph convolution operation~\cite{wang2019dynamic} is performed to obtain temporal and semantic contexts for snippets.
Zhao {\it et al.}~\cite{zhao2021video} introduce a video self-stitching graph network to address the scaling curse in \cite{xu2020g} by utilizing the graph network to exploit multi-level correlations among cross-scale snippets.
Besides temporal action localization, GNNs are utilized in \cite{bai2020boundary} and \cite{huang2020improving} for temporal action proposal generation and video action segmentation, respectively.

\subsection{Others}
\textbf{Video Object Segmentation} Wang {\it et al.}~\cite{wang2019zero} adopt GNNs to capture higher-order relations among video frames to help object segmentation from a global perspective.
In addition, they adapt the convolution operation in classical GNNs to the pixel-level segmentation task.
Lu {\it et al.}~\cite{lu2020video} extend~\cite{wang2019zero} by introducing a graph memory network to store memory in a graph structure with memory cells as nodes. And the query of a video frame in the memory graph facilitates segmentation mask prediction of that frame.

\noindent\textbf{Multi-Object Tracking} aims to track all the objects in a video. Two-step tracking-by-detection methods set up detection graphs where object detection instances in all frames are taken as nodes and an edge connects two instances in consecutive frames. The trajectory of an object forms a connected component in such a graph. Braso {\it et al.}~\cite{braso2020learning} propose a time-aware message passing network on the detection graph to encode its temporal structure and predict whether an edge is active or non-active to obtain the connected components. Weng {\it et al.}~\cite{weng2020gnn3dmot} suggest first connecting tracked objects and detected objects in two consecutive frames according to their spatial distances. Then, GNNs~\cite{wang2019dynamic} are used to aggregate features from neighbors and predict the status of edges to obtain the connectivity between the tracked objects and the detected ones.

\noindent\textbf{Human motion prediction} forecasts future poses conditioned on the past motions in videos. Like skeleton-based action recognition, spatio-temporal correlations among body parts are crucial cues for accurate motion prediction.
Li {\it et al.}~\cite{li2020dynamic} propose dynamic multi-scale GNNs to capture physical constraints and movement relations among body components at multiple scales and achieve effective motion prediction. Instead of modeling spatial correlation, Sofianos {\it et al.}~\cite{sofianos2021space} encode spatial joint-joint, temporal time-time, and spatio-temporal joint-time interactions via the proposed separable space-time GNNs. In addition to human motion prediction, Cai {\it et al.}~\cite{cai2019exploiting} exploit GNNs to estimate 3D human poses from consecutive 2D poses.

\noindent\textbf{Trajectory Prediction} aims to predict the future trajectory of a pedestrian on the basis of his/her existing trajectory and the complex interactions between the pedestrian and the environment/other pedestrians. Mohamed {\it et al.}~\cite{mohamed2020social} propose a graph representation of the pedestrian trajectory based on relative locations among pedestrians and their temporal information.
Furthermore, GCNs are applied to capture the complex interactions and the spatio-temporal graph representation.
Similar to \cite{mohamed2020social}, Yu {\it et al.}~\cite{yu2020spatio} propose a spatio-temporal graph Transformer to conduct crowd trajectory prediction according to relative locations among pedestrians and temporal dependencies among trajectories. In addition, they extend GATs by using Transformer's self-attention mechanism~\cite{vaswani2017attention}.
Beyond the distance between pedestrians,
Sun {\it et al.}~\cite{sun2020recursive} utilize social-related annotations, historical trajectories, and human contexts to construct a social behavior graph with pedestrians' historical trajectories as nodes and social relations as edges. Then, they build a GCN on the graph to learn higher-order social relations to facilitate future trajectory generation.
Shi {\it et al.}~\cite{shi2021sgcn} use the same GCNs in \cite{mohamed2020social} to capture spatial and temporal dependencies but they introduce two sparse directed spatial and temporal graph representations for trajectories. Different from previous methods,
Li {\it et al.}~\cite{li2022graph} construct a multi-scale graph to model the spatial information and surrounding area of pedestrians in multiple scales. In addition, they use scene semantic segmentation to help model the relations between pedestrians and the scene.

\subsection{Discussion}
Compared to images, videos involve temporal connectivity patterns among consecutive frames.
The spatio-temporal dependencies and contexts in a video play an essential role in video understanding.
Therefore, since the introduction of GNNs for capturing the spatial and semantic relations among visual constituents over time, video understanding has achieved significant progress.
However, existing methods usually either capture partial dependencies at the frame level or local contexts at the region level for several consecutive frames. A meaningful direction for future research would be effectively capturing long-range global dependencies or crucial contexts without redundant information.

\section{Vision + Language}
\label{sec:VL}
In addition to a single modality, 
there has been growing interest in applying GNNs to vision-and-language tasks, such as visual question answering~\cite{antol2015vqa}, visual grounding~\cite{kazemzadeh2014referitgame}, and image captioning~\cite{farhadi2010every}. GNNs facilitate modeling the structure of visual and linguistic components to help with cross-modal semantic alignment and joint understanding, which are mainly utilized to (i) learn the spatial or semantic relation between visual components, such as detected objects in images~\cite{teney2017graph}; (ii) model the dependencies among noun phrases in the input questions or expressions~\cite{bajaj2019g3raphground}; 
(iii) learn the relation among both images and texts jointly by combining the above two methods or performing graph convolution on the language-conditioned visual graph~\cite{yang2019cross}. 


\subsection{Visual Question Answering}
The input to visual question answering (VQA) is an image and a question expressed as text in a natural language, and the output is an answer to the question.
Since questions usually describe not only the appearances and attributes of the visual constituents of the image but also their relations, capturing the relations and aligning them with the question plays a vital role in VQA. 
Compared with conventional CNN-based or LSTM-based methods, 
GNN-based methods are able to capture complex object relationships, adaptively focus on relevant image parts, and conduct higher-order reasoning about the visual content. This enables more context-sensitive and detailed answers.
In VQA, graphs are constructed by leveraging the objects presented in image $\hat{I}$ and the textual content of question $\hat{Q}$. In the case of fact-based VQA, an additional graph is created to incorporate external knowledge $\hat{K}$. The general pipeline of GNN-based methods for the VQA task can be outlined as follows:
\begin{align}
    \small
        \mathcal{G} &= \Pi (\hat{I}, \hat{Q}, \hat{K}),\;
        \mathcal{F} = \mathtt{P}(\text{GNN}(\mathcal{G})),
    \label{eq:VQA}
\end{align}
where $\mathtt{P}(\cdot)$ refers to the final prediction layer, and $\mathcal{F}$ represents the final answer.

\noindent\textbf{Static Graphs.}
A typical solution sets up the graph representations of images and questions, and further aligns the constructed graphs~\cite{teney2017graph}. 
In the image graph, visual objects become nodes, and relative spatial relations between objects become edge weights; the question graph represents the dependencies among words via a dependency parser~\cite{de2008stanford}.
A recurrent unit~\cite{cho2014learning} followed by an attention mechanism similar to cosine similarity with learnable transformations is used to process and align the two graphs for prediction. 
Instead of the language-irrelevant image graph construction, Norcliffe {\it et al.}~\cite{norcliffe2018learning} use a graph learner to generate a sparse graph representation conditioned on the question for the image, followed by a graph CNN~\cite{monti2017geometric} to capture language-relevant spatial relations between objects. 

\begin{figure}[!t]
	\centering
	\includegraphics[width=\columnwidth]{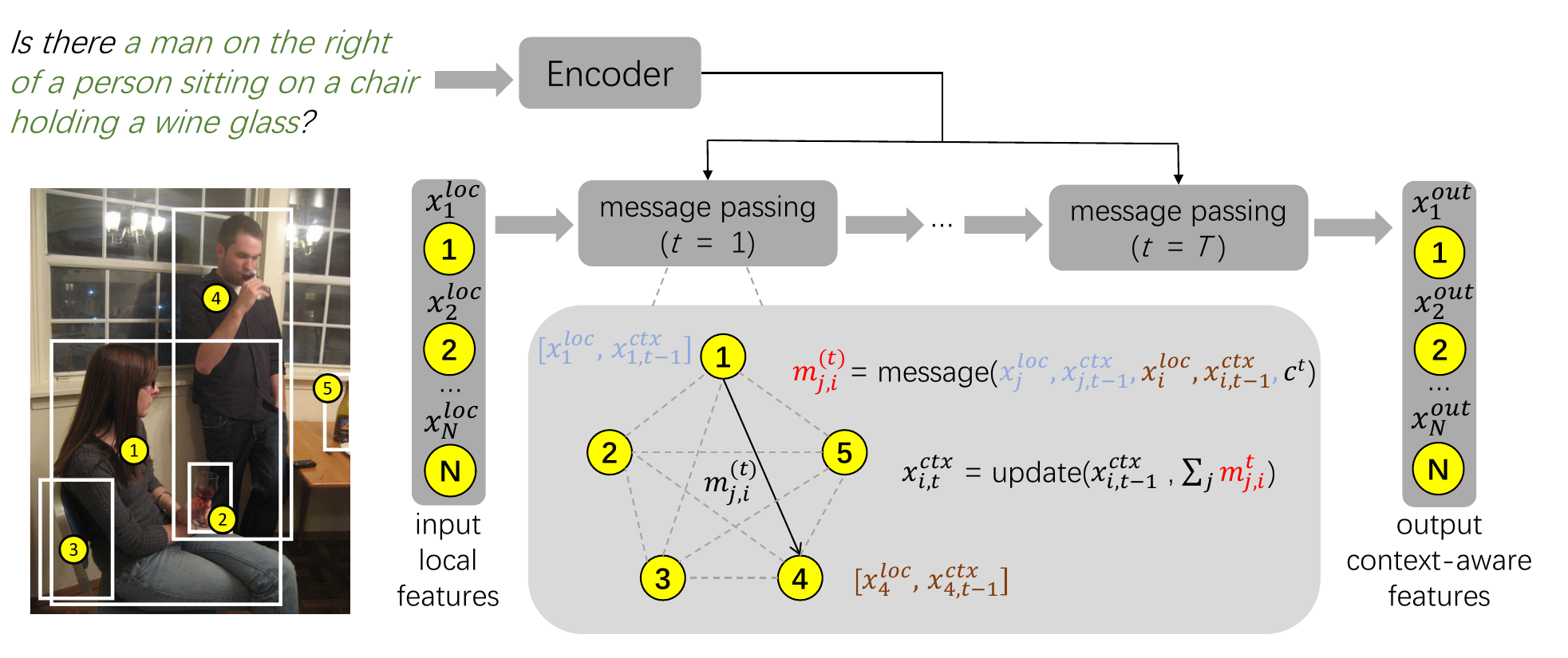}
	\caption{Language-Conditioned Graph Networks~\cite{hu2019language}.}
    \label{fig:vl134}
    \vspace{-0.2cm}
\end{figure}

\noindent\textbf{Dynamic Graphs.}
Instead of constructing static graphs, Hu {\it et al.}~\cite{hu2019language} generate a dynamic visual graph where the edges between objects are updated during each GNN-based message-passing iteration. The update is conditioned on the textual command vector extracted for each iteration (see Fig.~\ref{fig:vl134}). Jing {\it et al.}~\cite{jing2022maintaining} adopt a similar architecture to answer multiple questions for compositional VQA.
Li {\it et al.}~\cite{li2019relation} propose three graphs for representing semantic relations, spatial relations, and implicit relations between objects, respectively. A GAT is run on each graph to assign importance to nodes and learn relation-aware node representations.
Results from the GATs are aggregated to predict the answer.

\noindent\textbf{Symbolic Graphs.} 
Compared with feature-based graphs, symbolic graphs provide a clear and interpretable representation through symbolic labels, and excel in reasoning tasks.
Kim {\it et al.}~\cite{kim2020hypergraph} suggest using symbolic graphs generated through the scene graph parser~\cite{johnson2015image} and dependency parser, respectively, to represent the image and question.
Nodes of such symbolic graphs represent semantic units (\emph{e.g.,} attributes and objects) in a textual form, and edges represent relations (\emph{e.g.,} predicates) between semantic units.
The message passing neural network \cite{gilmer2017neural} is applied to the two symbolic graphs to obtain informative representations and align their sub-graphs.
Saqur {\it et al.}~\cite{saqur2020multimodal} 
further utilize the graph isomorphism network~\cite{xu2018powerful} to perform graph matching.

A comparison of these GNN-based methods on the VQA v2.0~\cite{goyal2017making} and the GQA~\cite{hudson2019gqa} datasets are shown in Tab.~\ref{tab:vl}.
In addition, GNNs have been applied to TextVQA~\cite{gao2020multi} and visual commonsense reasoning~\cite{wu2019connective, yu2019heterogeneous}, which extend the VQA task by reading texts in images and enabling commonsense reasoning, respectively.

\begin{table}
	\renewcommand\arraystretch{1.2}
	\centering
        \caption{Accuracy on two datasets for visual question answering. GNN-based methods are marked with $^\star$.}
        \vspace{-3mm}
	\resizebox{\linewidth}{!}{%
		\begin{tabular}{l|c|c} 
			\hline
			~~ Method~~            & \begin{tabular}[c]{@{}c@{}}~~~ VQA v2.0~\cite{goyal2017making}\\~~~~~ std ~ ~ ~ ~ ~~~ ~ dev ~ \end{tabular} & \begin{tabular}[c]{@{}c@{}}~~~ GQA~\cite{hudson2019gqa}\\~~~~~ val ~~~~~~~ test-dev ~ ~ ~ test ~ \end{tabular}  \\ 
			\hline
			Bottom-Up~\cite{teney2018tips}        & 65.67 ~ ~ ~~~ ~ ~~~ -                                                            & ~~ 52.2 ~ ~ ~ ~ ~~ -~ ~ ~ ~ ~ ~~ 49.7                                                        \\
			Graph Learner~\cite{norcliffe2018learning}$^\star$  & 66.18 ~ ~ ~~~ ~ ~~~ -                                                            & ~~ - ~ ~ ~ ~ ~ ~ ~ -~ ~ ~ ~ ~ ~ ~ ~ -                                                        \\
			single-hop + LCGN~\cite{hu2019language}$^\star$   &  ~~~~-~ ~ ~~~ ~ ~~~~~~~ -                                                                    & ~~ 63.9~~~~ ~ ~~ 55.8~ ~ ~~ ~~ 56.1                                                          \\
			Jing {\it et al.}~\cite{jing2022maintaining}$^\star$ & ~~~~-~ ~ ~~~ ~ ~~~~~~~ -                                                               & ~~~~ - ~ ~~ ~ ~~ ~ ~ - ~~ ~ ~ ~ ~~ 59.6                                                     \\
			ReGAT~\cite{li2019relation}$^\star$   & ~~~~~~~ 70.58~~ ~~ ~ ~~~ 70.27~~~~                                                 & ~~ - ~ ~ ~ ~ ~ ~ ~ -~ ~ ~ ~ ~ ~ ~ ~ -                                                        \\
			HANs~\cite{kim2020hypergraph}$^\star$     & ~~~~~~~ -~ ~  ~~ ~ ~~~~~ 69.10                                                      & ~~ - ~ ~ ~ ~ ~~~ 69.5~ ~ ~ ~ ~ ~~ -                                                            \\
			\hline
		\end{tabular}
	}
	\label{tab:vl}
	\vspace{-0.5cm}
\end{table}

\noindent\textbf{Fact-/Knowledge-based VQA.} 
Compared to classical VQA, fact-based VQA~\cite{wang2017fvqa} has an external knowledge base of facts to facilitate fact-based reasoning related to the image and question. 
A supporting fact is required to predict the correct answer for an image-question pair. 
Narasimhan {\it et al.}~\cite{narasimhan2018out} first parse and retrieve relevant facts from the external knowledge base using the image and question, and the retrieved facts are used to construct a fact graph. Then, they utilize a GCN to perform reasoning over the retrieved facts to identify the supporting fact and answer.
In addition to fact-based VQA, 
Singh {\it et al.}~\cite{singh2019strings} extend the classical gated GNN~\cite{li2015gated} to integrate information from the knowledge base, question, and image for knowledge-enabled VQA. They further use the knowledge base to interpret text in the image.
Marino {\it et al.}~\cite{marino2021krisp} construct a symbolic knowledge graph from the knowledge base and perform explicit reasoning on the graph via a relational GCN~\cite{schlichtkrull2018modeling}.
Then, the results of such explicit reasoning and Transformer-based implicit reasoning are integrated for answer prediction.

\subsection{Visual Grounding}
Visual grounding aims to locate the referent in an input image given a natural language expression. 
As in the VQA task, GNNs are used to capture dependencies in visual and linguistic components.
One type~\cite{yang2019dga} of GNN-based pipeline for visual grounding is similar to the one in VQA (Eq.~\ref{eq:VQA}) but replaces the prediction layer with a bounding box matching branch or bounding box refiner. Alternatively, other pipelines~\cite{bajaj2019g3raphground} involve employing a separate text encoder, followed by the fusion of textual information into the visual graph in the following manner:
\begin{align}
    \small
        \mathcal{G} = \Pi (\hat{I}),\;
        T = \mathtt{E}(\hat{T}),\;
        \mathcal{F} &= \mathtt{P}(\text{GNN}(\mathcal{G}, T)),
\end{align}
where $\mathtt{E}(\cdot)$ denotes the text encoder, and $T$ is the corresponding text embedding. 
GNNs encode structural information in both visual and linguistic data by effectively capturing the dependencies among distinct objects in two modalities, thus enhancing their ability to handle tasks involving diverse object interactions.

\begin{figure}[!t]
	\centering
	\includegraphics[width=\columnwidth]{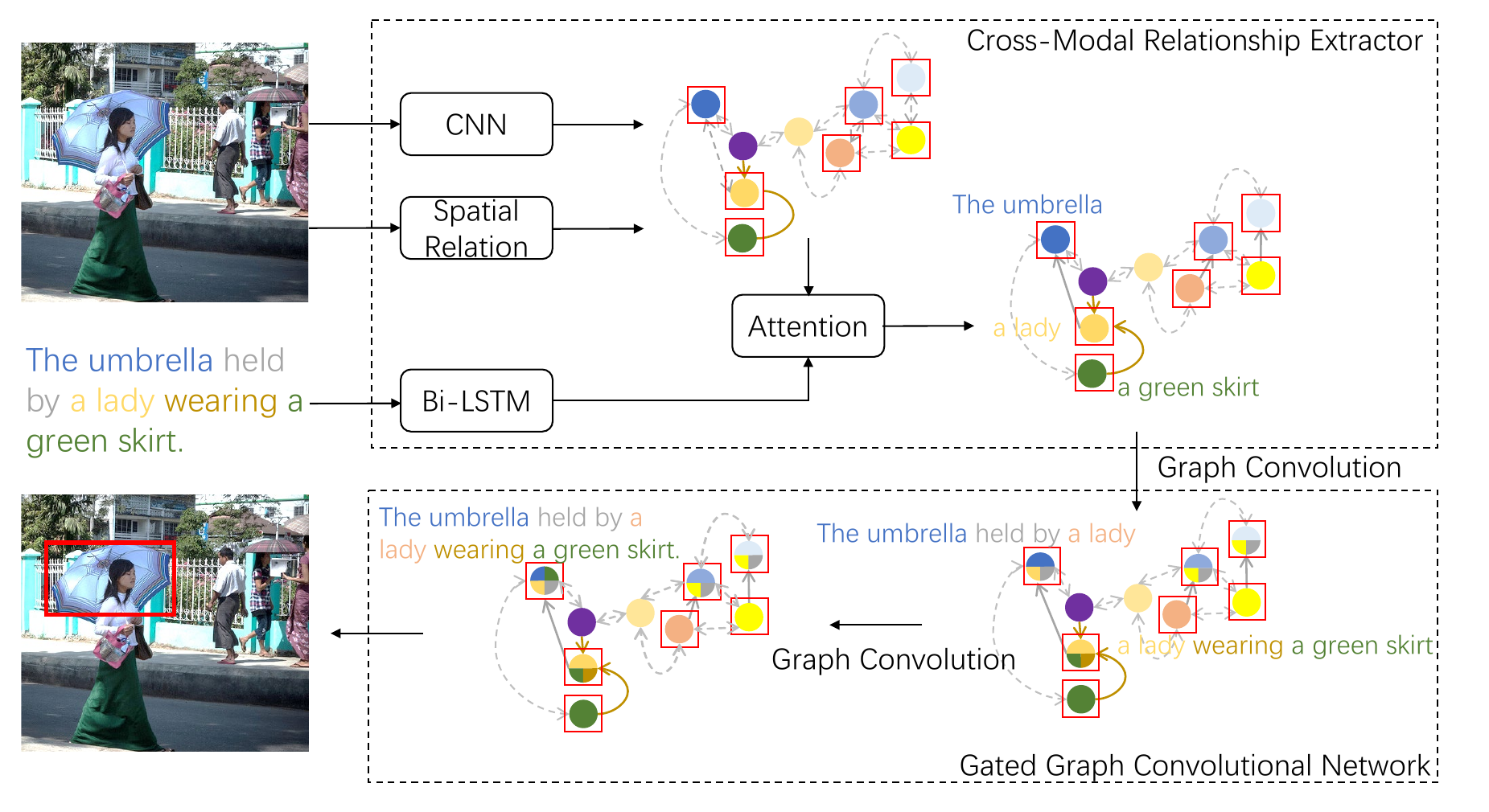}
	\caption{Cross-Modal Relationship Inference Network~\cite{yang2019cross}. Different colors correspond to different proposals after fusing with language. Then graph convolution is utilized to model the semantic contexts with multi-order relationships. The output is the proposal corresponding to the node most matching the language.} 
    \label{fig:vl155}
    \vspace{-0.2cm}
\end{figure}

\noindent\textbf{Referring Expression Grounding.} As shown in Fig.~\ref{fig:vl155}, Yang {\it et al.}~\cite{yang2019cross} construct a language-guided visual relation graph, that is conditioned on the expression, to represent the image. They apply a gated GCN to the graph to learn visual-linguistic fusion and capture multi-order semantic contexts. 
Yang {\it et al.}~\cite{yang2019dga} follow the same graph construction~\cite{yang2019cross}, but perform dynamic graph attention and convolution on the graph to achieve stepwise graph reasoning under the guidance of the linguistic structure of the expression. 
Similar to~\cite{yang2019dga}, Hu {\it et al.}~\cite{hu2019language} also generate a dynamic visual graph conditioned on the expression to capture complex relations. 
Instead of multi-order relations considered in \cite{yang2019cross,yang2020gre}, Wang {\it et al.}~\cite{wang2019neighbourhood} form a visual graph using direct intra-relations and inter-relations between neighboring objects. They use language-guided graph attention and aggregation to update node features. Different from previous methods, Chen {\it et al.}~\cite{chen2022multi} construct a multi-modal graph with randomly initialized bounding boxes as nodes and semantic relations between nearest neighboring nodes as edges. They utilize a GAT to update node features and bounding boxes, and introduce a graph Transformer to prune the nodes and edges iteratively. In addition, GNNs~\cite{kipf2017semi, wang2019dynamic} have been utilized for referring expression segmentation~\cite{huang2020referring, hui2020linguistic} and 3D visual grounding on point clouds~\cite{achlioptas2020referit3d} in a similar manner as for referring expression grounding.

\noindent\textbf{Phrase Grounding.} Phrase grounding locates the referent in the image for each phrase in the expression. The referent of one phrase may depend on the referents of other phrases. Bajaj {\it et al.}~\cite{bajaj2019g3raphground} first construct a phrase graph, where nodes correspond to noun phrases and edges connect pairs of noun phrases, and then apply a gated GNN~\cite{li2015gated} to capture the dependencies among phrases. They also perform similar graph computations on a visual graph and a visual-linguistic fusion graph to jointly predict the referent for each phrase. Unlike the approach in~\cite{bajaj2019g3raphground}, Yang {\it et al.}~\cite{yang2020propagating} perform dependency parsing on the expression to construct a language scene graph with noun phrases as nodes and relations between phrases as edges. They develop a one-stage graph-based propagation network for phrase grounding. The methods in \cite{liu2021relation, liu2020learning} use the same language scene graph in~\cite{yang2020propagating}, and apply GCNs and attention mechanisms to the graph to learn visual features for graph nodes. In addition to the language scene graph, the method in \cite{mu2021disentangled} constructs a visual scene graph~\cite{zellers2018neural} and extends the standard GCN to learn disentangled representations for various motifs in the graphs and capture motif-aware contexts.

\noindent\textbf{Video Grounding.} 
For video grounding, it is vital to capture temporal or spatial-temporal correlation. 
Zhang {\it et al.}~\cite{zhang2020does} construct a spatial relation graph over objects in each frame and a temporal dynamic graph where nodes correspond to objects in all frames and edges connect the same objects in consecutive frames. 
Zhao {\it et al.}~\cite{zhao2021cascaded} develop a different way of constructing a temporal graph, where each node represents an interval, and each direct edge represents the overlapping relation between two intervals. 

\subsection{Image and Video Captioning}
\textbf{Image Captioning.} Image captioning aims to generate a complete and natural description of an image. 
The relations between objects in the image are natural priors for image description. 
Since GNNs model data using graphs, enabling them to capture contextual information among objects in images to enhance contextual awareness, which empowers GNNs to generate more accurate and coherent captions in comparison to traditional methods.
Yao {\it et al.}~\cite{yao2018exploring} construct a semantic graph and a spatial graph to represent objects and their semantic and spatial relations. Then, they extend the GCN~\cite{kipf2017semi} on the graphs to obtain object representations with inherent visual relations, and use the attention LSTM \cite{anderson2018bottom} to decode the caption from such representations. Yao {\it et al.}~\cite{yao2019hierarchy} use the same graph construction and GCN as in~\cite{yao2018exploring} and extend single-level object relation modeling into three hierarchical levels including the image, regions and objects.
Unlike previous work only capturing visual relations, Yang {\it et al.}~\cite{yang2019auto} use a GCN and sentence reconstruction to capture the language inductive bias to form a dictionary at the training stage. At the inference stage, they utilize attention between an image scene graph and the dictionary to embed the language inductive bias into visual representations. Unlike previous methods directly decoding GNN-based context representations of the image scene graph into sentences, the method in \cite{zhong2020comprehensive} decomposes a scene graph into a set of sub-graphs and chooses important sub-graphs to decode at each time step. To achieve user desired captioning and improve the diversity at a fine-grained level, Chen {\it et al.}~\cite{chen2020say} use an abstract scene graph to control the described objects, attributes, and relations in the caption. A multi-relational GCN~\cite{schlichtkrull2018modeling} is used to encode the abstract scene graph grounded in the image. Nguyen {\it et al.}~\cite{nguyen2021defense} combine an image scene graph with a human-object interaction graph extracted from \cite{ulutan2020vsgnet} to enhance salient parts of the scene graph. Then, they use a similar GCN-LSTM architecture as in \cite{yao2018exploring} to generate the caption.

\noindent\textbf{Video Captioning.} Video captioning requires generating a description for a video. To learn the interactions among objects over time, Zhang {\it et al.}~\cite{zhang2020object} construct an object relation graph to connect each object with top-k corresponding objects in all frames and use a GCN to learn the relations on the graph. As for image captioning, they generate descriptions corresponding to object relational features via an attention LSTM~\cite{anderson2018bottom}. With the same motivation as~\cite{zhang2020object}, Pan {\it et al.}~\cite{pan2020spatio} construct a spatio-temporal graph and adopt a GCN to update node features. For graph construction, they first take objects in individual frames as nodes and normalized IOU between objects as edge weights, then connect semantically similar objects in two consecutive frames.
Contrary to previous GCN-LSTM architectures, Chen {\it et al.}~\cite{chen2021motion} aggregate information in a temporal graph and feed the aggregation result into an LSTM decoder, which adjusts the graph structure according to its hidden state and, in the meantime, updates the hidden state according to the graph representation.
\vspace{-0.2cm}
\subsection{Others}
\noindent\textbf{Image-Text Matching.}
Image-text matching aims to measure semantic similarity between a pair of image and sentence, which plays a vital role in visuo-linguistic cross-modal content retrieval. Since local similarity between regions and words contributes to global semantic similarity, Huang {\it et al.}~\cite{huang2019acmm} utilize a cross-modal GCN to align visual regions and word representations for global similarity computation. Unlike the method in \cite{huang2019acmm}, Li {\it et al.}~\cite{li2019visual} first learn a relation-enhanced global image representation based on local regions and their relations, and then match this image representation with the sentence representation. A GCN is used to perform relation reasoning on image regions to generate relation-enhanced features. Similar to the method in~\cite{li2019visual}, Yan {\it et al.}~\cite{yan2021discrete} also use a GCN to reason about semantic relations between image regions. In addition, they propose a discrete-continuous policy gradient algorithm to transform images and texts into a common space. Wang {\it et al.}~\cite{wang2020consensus} improve region-based relation reasoning by introducing commonsense knowledge into the reasoning process. They construct a concept correlation graph and learn consensus-aware concept representations via stacked GCN layers. Besides constructing a relation graph on image regions, Liu {\it et al.}~\cite{liu2020graph} and Li {\it et al.}~\cite{li2020visual} build a textual graph for the sentence via a dependency parser and perform semantic matching between the two graphs via a GCN and an attention mechanism. In addition, GNNs have also been utilized for video-text matching~\cite{chen2020fine, zeng2021multi, li2021adaptive}.


\noindent\textbf{Vision-Language Navigation.} Vision-language navigation requires an agent to navigate an unknown environment following a natural language instruction~\cite{anderson2018vision}. Deng {\it et al.}~\cite{deng2020evolving} develop an evolving graphical planner model, which iteratively and dynamically expands a trajectory graph built on the current node, visited nodes, potential action nodes and their relations, and runs a GNN on the graph to predict an action. Unlike the method in~\cite{deng2020evolving}, Chen {\it et al.}~\cite{chen2021topological} use a GNN to capture the visual appearance and connectivity of the environment from its topological map. Gao {\it et al.}~\cite{gao2021room} involve external knowledge in action prediction and run a GCN on external and internal environment knowledge graphs to extract external knowledge used for cooperating with internal knowledge. Chen {\it et al.}~\cite{chen2022reinforced} generate a layout graph at each step according to the instruction and use graph convolution to learn the current navigation state from the current layout graph and history layout memory. The navigation state is used to predict the action probability distribution at the current step. Like the method in \cite{deng2020evolving}, Chen {\it et al.}~\cite{chen2022think} construct an environment graph on visited, navigable, and current nodes. Furthermore, they aggregate information in the graph using self-attention to predict the navigation score for each node.

\noindent\textbf{Video Question Answering.}
Similar to the VQA approach in \cite{teney2017graph}, Park {\it et al.}~\cite{park2021bridge} represent the question as a parse question graph.
To capture temporal interactions in a video,
they also construct a frame-based appearance graph and a motion graph.
Then, GCNs~\cite{kipf2017semi} are used to learn question-appearance and question-motion interactions on the three graphs.
Unlike previous methods that use GNNs to learn relation-embedded node representations,
Yu {\it et al.}~\cite{yu2021learning} utilize GNNs to adaptively propagate predicted probabilities among samples in the video to obtain consistent results for samples with high visual similarity. Instead of exploiting frame or motion-level information, some works \cite{liu2021hair, huang2020location} suggest constructing object-level visual graphs.
In particular, Liu {\it et al.}~\cite{liu2021hair} combine GNNs and memory networks~\cite{sukhbaatar2015end} to perform dynamic relation reasoning on visual and semantic graphs at the object level.

\subsection{Discussion}
Natural language sentences contain natural dependency structures, while images can take the form of spatial, semantic, or scene graph representations. GNNs thus are utilized naturally by the vision-and-language community to capture the relations among visual, linguistic, or multi-modal components, or perform cross-modal alignment between visual and linguistic graphs. Nevertheless, there still exist open problems in the exploitation of GNNs in vision-and-language research. For example, is it possible to design a unified GNN-based architecture for different vision-and-language tasks, especially in view of the similar roles played by GNNs in different tasks? How do we better explore the reasoning ability and explainability of GNNs beyond neighborhood aggregation and message passing? How do we further incorporate knowledge graphs for achieving knowledge-aware graph reasoning?

\section{3D Data Analysis}
\subsection{3D Representation Learning}
Among various 3D data representations, point clouds and meshes receive increasing attention due to their strong ability to represent irregular structures and complicated 3D shapes.
In this task, GNNs can efficiently capture complex geometric and topological relationships inherent in 3D structures and are versatile in handling varying sizes and resolutions of 3D data, making them ideal for tasks that require robust and scalable 3D analysis.




\subsubsection{Point Cloud Representation}
PointNet~\cite{qi2017pointnet} is a pioneering work for point cloud representation learning that takes the raw point cloud as the input and processes each point independently for feature representation. Such point-based methods include PointNet++~\cite{qi2017pointnet++}, PointCNN~\cite{pointcnn}, etc. Another type of methods is geometry-based, including KPConv~\cite{kpconv} and MinkowskiNet~\cite{choy20194d}, which use mathematical models to represent the shape and structure of the point cloud. However, both point-based and geometry-based methods treat each point as an individual entity yet do not directly consider the relationships between points, which limits robustness when applied to more noisy, complex, or incomplete point clouds. 

\begin{figure}[!t]
	\centering
	\includegraphics[width=0.46\textwidth]{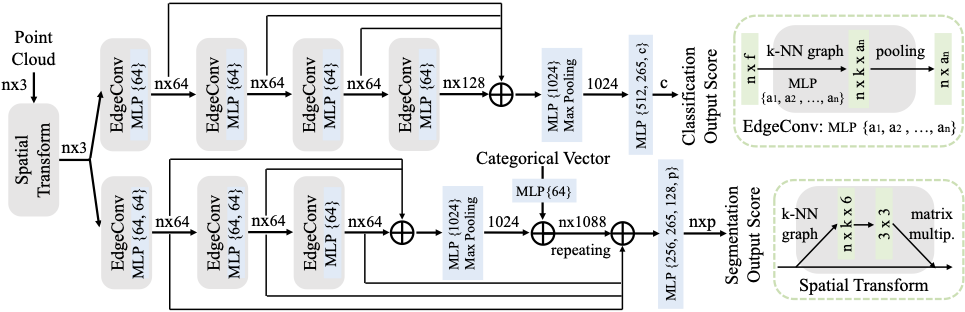}
	\caption{Dynamic Graph Convolution Neural Network (Figure used courtesy of~\cite{wang2019dynamic}).}\label{fig:DGCNN}
    \vspace{-0.5cm}
\end{figure}

Naturally, GNN-based approaches model the input point cloud as a spatial graph and apply a GNN or graph Transformer to characterize local or global interaction among points.
A pioneering graph-based approach, ECC~\cite{simonovsky2017dynamic}, generalizes convolutions on regular grids to point cloud based graphs. ECC computes adaptive convolution kernel weights using specific edge labels in the neighborhood of a vertex to better utilize edge information. A recent work, AdaptConv~\cite{zhou2021adaptive}, also focuses on generating adaptive kernels but according to learned features.
In the above methods, the graph structure is determined by the input point cloud, more specifically, local point neighborhoods, which limits flexibility and non-local relation modeling. To address these issues, DGCNN~\cite{wang2019dynamic} (cf. Fig.~\ref{fig:DGCNN}) designs an EdgeConv operator and proposes to construct graphs in the feature space and dynamically update graphs in each layer so that both global and local interactions can be flexibly captured. 
Following ECC and DGCNN, a series of works are contributed to improving the spatial-domain graph convolutions for point cloud representation, including \cite{zhang2019linked, shen2018mining, Hassani_2019_ICCV, chen2019clusternet, lin2020convolution, lei2020spherical}
Different from the mainstream approach, a random walk based method, CurveNet~\cite{xiang2021walk}, generates sequences of point segments as non-local descriptors to better depict point cloud geometry.

In contrast to performing graph processing operations in the spatial domain, spectral methods build on a mathematically elegant approach that defines graph convolution as spectral filtering, which relies on the eigen-decomposition of the graph Laplacian~\cite{li2018adaptive, syncspec, te2018rgcnn, pan20183dti, zhang2018graph}.
LocalSpecGCN~\cite{wang2018local} performs spectral filtering on dynamically built local graphs while recursively clustering spectral coordinates to support graph pooling, which improves PointNet++~\cite{qi2017pointnet++} via alleviating point isolation, and also addresses the limitation of traditional spectral GCNs that the graph Laplacian and pooling hierarchy need to be pre-computed for the full graph.
HGNN~\cite{feng2019hypergraph} designs a hyperedge convolution operation to learn high-order data correlation for representing the complex structure in a point cloud. Instead of using a fixed view, DiffConv~\cite{diffconv} is operated under adaptive views to match local point cloud density. 
Wiersma~\textit{et al.}~\cite{DeltaConv} design a graph-based anistropic convolutional operator by combining a set of geometric operators defined on scalar and vector fields to encode the directional information of each surface point.


In addition to graph convolution based methods, another line of work aims to improve graph attention neural networks for point cloud analysis \cite{chen2019gapnet, wang2019graph, xun2021graph, huang2022dual}. In such work, during graph feature aggregation, attention weights are computed from both neighbors' positions and learnable features to adaptively attend important graph nodes or regions. In addition, graph-RNN is explored in \cite{gomes2021spatio} to consider relations between both spatially and temporally neighboring points for point cloud sequence analysis.

As Transformers have been extensively explored in computer vision, graph-based Transformers designed for 3D point cloud representation learning have also been proposed~\cite{zhao2021point, guo2021pct}. The core mechanism of Transformers, self-attention, naturally enables the network to build and represent global relations. PT~\cite{zhao2021point} computes self-attention on local k-NN graphs, which leads to high-quality local representations while global relations are captured in deep layers. In comparison, PCT~\cite{guo2021pct} adapts global self-attention to learn both local and non-local interactions, which sacrifices computational efficiency for representation capability.
Recently, Lu \textit{et al.}~\cite{lu20223dctn} further proposed 3DCTN to combine a Transformer and graph convolutions to achieve both efficient and powerful 3D point representations. 
As the agglomeration process of point clouds (including point sampling, grouping, and pooling) is complex, some methods \cite{liu2019dynamic, xu2020grid, lei2020spherical, zhang2021point, li2021towards} have been proposed for simplifying the procedure or promoting the efficiency.
The performance of some GNN-based algorithms on the point cloud classification task is given in Tab.~\ref{tab:3D-classification} for a comparison with non-GNN-based ones~\cite{qi2017pointnet,li2018pointcnn, kpconv}. On the ModelNet40 dataset, advanced GNN-based methods \cite{xiang2021walk, mohammadi2021pointview, zhao2021point, yan2022let} achieve SOTA performance of point cloud classification.
\vspace{-5mm}

\begin{center}
 \begin{table}[!t]
  \caption{\small Overall Accuracy (oAcc.) on the point cloud classification task of the ModelNet40 dataset.}
  \label{tab:3D-classification}
  \vspace{-0.2cm}
  \centering
  \setlength\tabcolsep{5pt}
  \scalebox{0.85}{
  \begin{tabular}{l|ll|c}
   \toprule
   & Method & Reference &  oAcc. (\%)  \\
   \midrule
                 & PointNet~\cite{qi2017pointnet} & CVPR'17 & 89.2 \\
   non-GNN-based & PointCNN~\cite{li2018pointcnn} & NeurIPS'18 & 92.2 \\
                 & KPConv~\cite{kpconv} & ICCV'19 & 92.5 \\
   \midrule
             & LocalSpecGCN~\cite{wang2018local} & ECCV'18 & 92.1 \\
             & DGCNN~\cite{dgcnn} & TOG'19 & 92.9 \\
   GNN-based & DeepGCN~\cite{li2019deepgcns} & ICCV'19 & 93.6 \\
             & PT~\cite{zhao2021point} & ICCV'21 & 93.7 \\
             & CurveNet~\cite{xiang2021walk} & ICCV'21 & 94.2 \\
   \bottomrule
  \end{tabular}}
  \vspace{-5mm}
 \end{table}
\end{center}

\subsubsection{Mesh Representation}
A polygonal mesh discretely represents the surface of a 3D shape with faces and vertices. A mesh has a set of vertices that are connected by edges to form faces, therefore, a mesh can be directly represented as an undirected graph.

\begin{figure}[!t]
	\centering
	\includegraphics[width=0.46\textwidth]{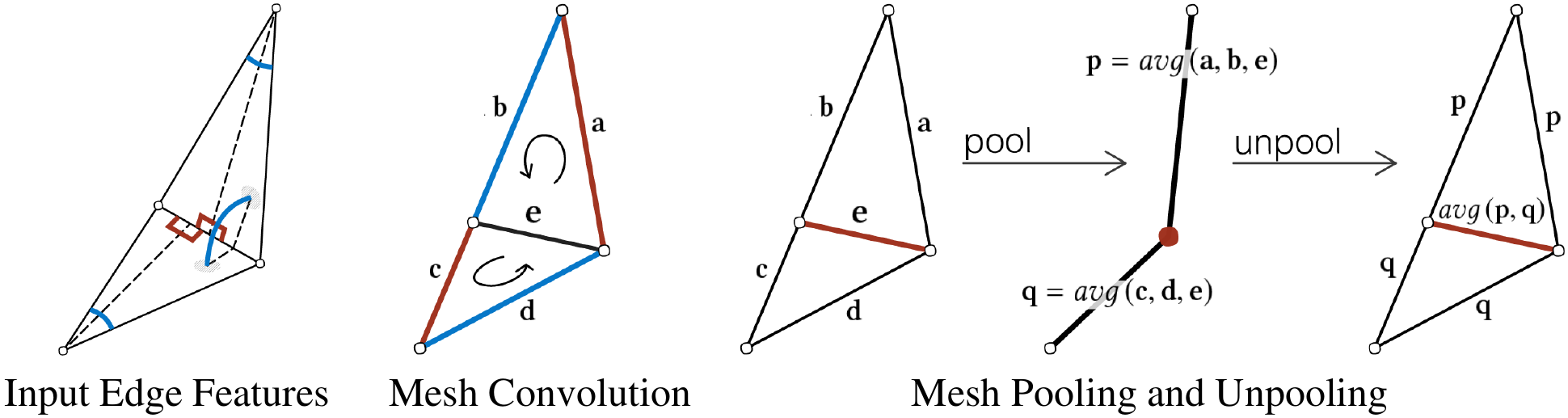}
	\caption{MeshCNN (Figure used courtesy of~\cite{hanocka2019meshcnn}).}\label{fig:MeshCNN}
    \vspace{-0.3cm}
\end{figure}

A line of work explores representation learning over meshes by performing operations such as convolution and pooling on different mesh components, \textit{e.g.,} vertices~\cite{verma2018feastnet}, edges~\cite{hanocka2019meshcnn}, faces~\cite{feng2019meshnet}, or edges and vertices~\cite{schult2020dualconvmesh}. FeaStNet~\cite{verma2018feastnet} performs graph convolution in each local neighborhood centered at a vertex to aggregate the features at vertices connected to the center vertex. As local neighborhoods in a mesh have irregular structures, FeaStNet learns multiple weight matrices for graph convolution and dynamically chooses one of the weight matrices for each vertex in a neighborhood based on the vertex feature.
MeshCNN~\cite{hanocka2019meshcnn} adopts a different approach, where mesh convolution, pooling, and unpooling are operated on edges since every edge is adjacent to two triangle faces defined by additional four nearby edges, as illustrated in Fig.~\ref{fig:MeshCNN}. Such edge-oriented operations have a fixed neighborhood size and derive a simple network for mesh feature extraction. Besides, MeshNet~\cite{feng2019meshnet} introduces convolution operated on mesh facets as well as feature splitting to obtain spatial and structural mesh features. DCM-Net~\cite{schult2020dualconvmesh} performs graph-based operations on both edges and vertices by jointly exploiting geodesic and Euclidean convolutions, which encourages contextual information flow among spatially or geodesically nearby patches. PD-MeshNet~\cite{milano2020primal} constructs a pair of primal and dual graphs on an input 3D mesh to connect the features of mesh faces and vertices and employs a graph attention network to dynamically aggregate these features.

Recently, many studies have focused on designing new GNN operators to improve feature extraction.
SpiralNet++~\cite{gong2019spiralnet++} improves \cite{bouritsas2019neural} with a fast and efficient intrinsic mesh convolution operator that fuses features from adjacent nodes with multi-scale local geometric information. 
A stacked dilated mesh convolution is proposed to inflate the receptive field of graph convolution kernels~\cite{singh2021mesh}. PolyNet~\cite{yavartanoo2021polynet} develops convolution and pooling operations that are invariant to the scale, size, and perturbations of local patches. 
He~\etal~\cite{curvanet} improve the graph convolution operation by learning direction sensitive geometric features from mesh surfaces. 
GET~\cite{he2021gauge} 
builds an efficient Transformer with both gauge equivariance and rotation invariance on triangle meshes. 
Recently, SubdivNet~\cite{hu2022subdivision} performs representation learning on meshes using connectivity from the Loop subdivision sequence, by building hierarchical mesh pyramids.
Dong~\etal~\cite{dong2022laplacian2mesh} consider graph convolutions and pooling in the spectral domain, and maps mesh features in the Euclidean space to the multi-dimensional Laplacian-Beltrami space. 
Different from the custom of constructing graphs with regular neighborhood structures, 
MeshWalker~\cite{lahav2020meshwalker} takes a similar approach as in \cite{xiang2021walk} for mesh analysis, and random walks along edges are used for 3D shape information extraction. AttWalk~\cite{attwalk} enhances~\cite{lahav2020meshwalker} by exploring meaningful interactions among different walks. 

\subsection{3D Understanding}
GNNs are powerful neural networks capable of representing both local and global regions, which leads to accurate context-aware semantic understanding. 3D understanding tasks that can exploit GNN algorithms include 3D segmentation, detection, and visual grounding.
Based on GNN-based 3D representation learning approaches, the methods proposed for 3D understanding tasks further consider instance-level and/or scene-level features, and GNNs play an important role in learning inter-instance and contextual information. 
Denote the 3D input as $I$, 
a graph $\mathcal{G}$ is constructed first at the input level, and the GNN-based methods serve to extract low-level features $\mathcal{F}$ from $\mathcal{G}$. 
In some methods, the features $\mathcal{F}$ are further aggregated via constructing a region-level or instance-level graph $\mathcal{G'}$, on which GNN-based methods are employed to incorporate the high-level relations into a more representative feature $\mathcal{F'}$ for the downstream task, such as segmentation or detection. The whole process can be formulated as: 
\begin{align}\label{eq:3Dunder}
    \footnotesize
    \mathcal{F} = \text{GNN}(\Pi(I)),\; \mathcal{F'} = \text{GNN}(\Pi(\mathcal{F})).
\end{align}

\begin{center}
 \begin{table}[!t]
  \caption{Performance (mIoU) on the point cloud semantic segmentation task of the S3DIS Area-5 dataset.}
  \label{tab:3D-segmentation}
  \vspace{-0.2cm}
  \centering
  \footnotesize
  \setlength\tabcolsep{5pt}
  \begin{tabular}{l|ll|c}
   \toprule
   & Method & Reference &  mIoU (\%)  \\
   \midrule
                 & PointNet~\cite{qi2017pointnet} & CVPR'17 & 41.1 \\
   non-GNN-based & PointCNN~\cite{li2018pointcnn} & NeurIPS'18 & 57.3 \\
                 & KPConv~\cite{kpconv} & ICCV'19 & 67.1 \\
   \midrule
             & SPG~\cite{superpointv2} & CVPR'19 & 61.7 \\
             & Point-Edge~\cite{pointedge} & ICCV'19 & 61.9 \\
   GNN-based & GACNet~\cite{wang2019graph} & CVPR'19 & 62.9 \\
             & PT~\cite{zhao2021point} & ICCV'21 & 70.4 \\
             & PTv2~\cite{zhao2021pointv2} & NeurIPS'22 & 72.6 \\
   \bottomrule
  \end{tabular}
  \vspace{-0.5cm}
 \end{table}
\end{center}

\vspace{-0.9cm}
\subsubsection{3D Point Cloud Segmentation}
3D segmentation, \emph{e.g.,} semantic segmentation, 
needs to understand both global scene semantics and fine-grained point-wise signatures. 
GNNs can efficiently capture complex inter-point, inter-region, and inter-object relationships, enabling robust feature extraction and scene understanding that adapt to irregular geometric structures.
Existing work primarily focuses on how to construct graphs and achieve better local feature aggregation and global interaction. Note that some studies may involve both aspects, but we categorize them based on their core contributions.



\noindent\textbf{Graph Construction.}
Qi \textit{et al.}~\cite{qi3d} make the first attempt to apply GCNs to this task. 
They construct a large-scale graph at the point level for each 3D scene to incorporate geometric relations in a learning-based way, so that semantic information from the RGB images and geometric information from the point clouds can be combined through a GNN.
Differently, SPG~\cite{superpoint} constructs graphs at the superpoint level. Each superpoint is a cluster of nearby points that have similar semantic meanings. 
A graph constructed on such geometric partitions has a simpler structure and smaller scale, giving rise to high-quality and efficient segmentation. 
Landrieu \textit{et al.}~\cite{superpointv2} further propose a graph-structured contrastive loss and a cross-partition weighting strategy to better oversegment a point cloud into superpoints in a supervised manner, which results in high contrast at object boundaries and a compact representation for each segment.

\noindent\textbf{Feature Aggregation.} To better handle irregular structures of point clouds, GACNet~\cite{gacnet} performs graph attention convolutions on local point subsets, where attention weights are learned from both 3D coordinates and point features, so that the network can attend most relevant neighboring points to better model the 3D structure of a point cloud.
To utilize all available local contextual information, PointWeb~\cite{pointweb} densely connects all point pairs in a local region and updates point features using nonlinearly transformed pairwise feature differences. 
Compared to focusing each local graph at a point, PointWeb's feature aggregation provides an improved depiction of local regions via enhanced interactions among adjacent points.
Jiang \textit{et al.}~\cite{pointedge} construct a hierarchical graph, where a point branch and an edge branch interact with each other.
PyramNet is equipped with a Graph Embedding Module~\cite{pyramnet} to strengthen its ability to represent local geometric details.
Combining depthwise graph convolutions and pointwise convolutions, HDGCN~\cite{8794052} customizes DGConv to better extract local features.
Lei \textit{et al.}~\cite{seggcn} build each local graph as a ball space around a center point, so that the weights of its neighboring points can be efficiently determined for convolution by its location pre-defined by a volumetric division in the ball. 
Grid-GCN~\cite{gridgcn} leverages the efficiency of a grid space to significantly accelerate the process of structuring point clouds.
Point2Node~\cite{point2node} explores the correlation among graph nodes at different scales.
RGCNN~\cite{te2018rgcnn} applies graph convolution in the spectral domain to part segmentation.
SegGroup~\cite{tao2022seggroup} runs GCNs to decrease the difference among nodes from the same instance and increase the variance among nodes from different instances.

The performance of GNN-based methods and non-GNN-based ones~\cite{qi2017pointnet, li2018pointcnn, engelmann2020dilated} on the point cloud semantic segmentation task is given in Tab.~\ref{tab:3D-segmentation}. As shown, on the S3DIS-Area5 dataset, point-transformer-based methods \cite{zhao2021point, zhao2021pointv2} achieve SOTA segmentation accuracy.

\subsubsection{3D Object Detection}
3D object detection aims to localize and recognize objects in point clouds or RGBD data. 
Compared with traditional 3D object detection methods, GNN-based methods improve feature extraction at the point level by capturing cross-point relations to facilitate non-local learning. 
Liang \textit{et al.}~\cite{continuous_detection} focused on multi-sensor detection by fusing image information from cameras with point clouds from LiDAR, and proposed continuous fusion layers, which compute dynamic convolution kernels for better integration between different modalities~\cite{deep_para}. 
Shi \textit{et al.}~\cite{Point-GNN} proposed Point-GNN for one-stage 3D object detection, where neighboring points within a fixed radius are connected, and an auto-registration method is designed to reduce the translation variance by dynamically refining point positions. 
HGNet~\cite{hgnn} adopts a GCN with a hierarchical structure to adequately exploit multi-level semantics for object detection. 

GNNs are also employed to capture inter-object relations through object-level graphs, which have been proven beneficial to the understanding of 3D scenes with multiple objects.
Feng \textit{et al.}~\cite{9234727} built an inter-object relation graph to enhance proposal features for higher-quality detection. 
Object DGCNN~\cite{wang2021object} models object relations via dynamic graph convolutions~\cite{dgcnn} to learn better features for object queries. 
PointRGCN~\cite{pointrgcn} introduces both point- and object-level GNNs to thoroughly model inter-point and inter-object relations as well as contextual information. 

\vspace{1mm}
\noindent
\textbf{Discussion.} GCN-based methods are the most popular solutions for 3D understanding, where both local neighborhoods and global contexts can be modeled. Recent works~\cite{guo2021pct, zhao2021point} proposed Transformer architectures for stronger feature extraction via the self-attention mechanism but at the cost of efficiency. However, the problem of a good combination of graph convolution and self-attention for 3D understanding has been less explored. 
A future direction would be a comprehensive framework that performs point-level, object-level, and scene-level understanding, and leverages the complementary advantages of MLP, graph convolution, and self-attention to balance between efficiency and performance.
In addition to point clouds and meshes, automated panoptic symbol spotting~\cite{fan2021floorplancad,fan2022cadtransformer}, which is crucial for creating 3D prototypes in architecture, engineering, and construction industries, introduces graphical networks to model symbol-wise dependencies and has received growing attention.
\subsection{3D Generation}
Approaches for 3D generation usually use an encoder for input representation and a decoder to transform latent features into 3D outputs. 
Besides being used in the encoder for feature extraction as in 3D understanding tasks (Eq.~\ref{eq:3Dunder}), 
GNNs are widely adopted to (i) learn required manipulations of a 3D input, \textit{e.g.}, adjusting point positions for the task of point cloud denoising; 
(ii) generate a 3D shape from a reference input, \textit{i.e.}, 3D reconstruction. 



\subsubsection{Point Cloud Completion and Upsampling}
Scanned data from 3D sensors are usually sparse, non-uniform, and incomplete. 
Point cloud completion and upsampling aim to produce complete and dense point clouds from real scans. 
Denote the sparse or incomplete input as $I$, it aims to output more points $\triangle I$ to obtain a dense or complete one $I' = \{I, \triangle I\}$. A typical pipeline uses a feature extractor and an upsampling network as follows:
\begin{equation}
	\footnotesize
        \mathcal{F'} = \mathtt{Feature}(I),\;
        I' = \mathtt{Upsample}(\mathcal{F'}).
    \label{eq:3Ddenoise}
\end{equation}
In this task, GNNs can efficiently and dynamically capture the geometric structure of point clouds, providing both global and local topological information to avoid generating inconsistent content.

The challenge of completion lies in recovering complete object shapes from incomplete information. 
ECG~\cite{pan2020ecg} is a two-stage framework that exploits skeleton as the intermediate structural representation and introduces a hierarchical encoder based on graph convolution to extract multi-scale edge features. 
 In \cite{zhu2021towards}, leap-type EdgeConv~\cite{wang2019dynamic} is used to capture local geometric features, and a cross-cascade module is used to hierarchically combine local and global features. 
 Shi~\emph{et al.}~\cite{shi2021graph} regard point cloud completion as a generation task, and cast the input data and intermediate generation as controlling and supporting points, and a GCN is utilized to guide the optimization process.
 
For the task of upsampling, a multi-step point cloud upsampling network is proposed in \cite{yifan2019patch} 
to compute the k-NN graph for feature aggregation. 
Unlike PU-GCN~\cite{qian2021pu} that integrates densely connected graph convolutions into an Inception module, AR-GCN~\cite{wu2019point} incorporates residual connections into a GCN to exploit the correlation between point clouds with different resolutions, as well as a graph adversarial loss to capture characteristics of high-resolution ones. PUGeo-Net~\cite{qian2020pugeo} extends DGCNN by two modules for feature recalibration and point expansion.

\subsubsection{3D Data Denoising}
Denote a 3D input with noise as $I$, the denoising task aims to estimate the noise component $N$, which leads to the denoised one $I' = I - N$. A typical pipeline uses a feature extractor and a denoising network:
\begin{equation}
	\footnotesize
        \mathcal{F'} = \mathtt{Feature}(I),\;
        I' = \mathtt{Denoise}(\mathcal{F'}),
    \label{eq:denoise}
\end{equation}
For this task, GNNs can adaptively learn from local neighborhood structures for noise reduction, are robust to irregular sampling, and integrate multiple features (\emph{e.g.,} position, normal, color) of each point in the data.

Given that the denoising task is more concerned with local representations of point neighborhoods, 
Pistilli~\emph{et al.}~\cite{pistilli2020learning} introduce a new dynamic graph framework to achieve better feature aggregation.
To address this issue,
DMR~\cite{luo2020differentiable} builds on DGCNN~\cite{wang2019dynamic} to learn the underlying manifold of the noisy input from differentiably subsampled points and their local features for less perturbation. GPDNet~\cite{pistilli2020learning} proposes to build hierarchies of local and non-local features to regularize the underlying noise in the input point cloud.
For 3D mesh denoising,
Armando \textit{et al.} develop a multi-scale GCN in \cite{armando2020mesh}, where the algorithm is built on CNN-based image denoising techniques. 
More recently, GCN-Denoiser~\cite{shen2022gcn} learns a rotation-invariant graph representation for local surface patches, and performs graph convolutions over both static graph structures of local patches and dynamic learnable structures. 
GeoBi-GNN~\cite{zhang2022geobi} excavates the dual-graph structure in meshes to capture both position and normal noises through a GNN-based U-Net.


\subsubsection{3D Reconstruction}
3D reconstruction recovers 3D point clouds or meshes from lower-dimensional inputs. 
Mainstream paradigm adopts an encoder-decoder architecture, where a low-dimensional input $I$ is encoded into latent features and then decoded into a 3D output:
\begin{equation}
	\footnotesize
        \mathcal{F'} = \mathtt{Encode}(I),\; 
        O  = \mathtt{Decode}(\mathcal{F'}), 
    \label{eq:3Drecon}
\end{equation}
For unconditional generation, only $\mathtt{Decode}$ is needed to decode a random code sampled from a pre-defined distribution over 3D outputs.
Compared with other methods, GNNs can provide sparse dependency modeling and more detailed geometric description for 3D reconstruction. 

\noindent
\textbf{Point Cloud Reconstruction.}
FoldingNet~\cite{yang2018foldingnet} has a graph-based auto-encoder that deforms a 2D grid into a point cloud with delicate structures.
In \cite{mo2019structurenet}, a unified latent space is defined for graph representations, and a GNN named StructureNet is able to encode or generate point clouds. 
Inspired by the success of GANs, some works focus on unsupervised point cloud generation, which incorporates graph convolution into the decoder~\cite{valsesia2018learning, shu20193d} or even the self-attention mechanism~\cite{li2021hsgan} for better generation quality. 

\noindent
\textbf{Mesh Reconstruction.} Pixel2Mesh~\cite{wang2018pixel2mesh} employs a GNN for mesh reconstruction from a single image. Instead of learning a direct mapping from images to meshes, targeted at multi-view mesh reconstruction, Pixel2Mesh++~\cite{wen2019pixel2mesh++} uses a GCN to predict a series of deformations to refine a coarsely generated shape.
Mesh R-CNN~\cite{gkioxari2019mesh} considers reconstruction from images with multiple objects.
It augments Mask R-CNN~\cite{he2017mask} with a GCN, which performs 3D shape inference by refining a coarsely cubified mesh.
For the task of mesh reconstruction from image volumes, Voxel2Mesh~\cite{wickramasinghe2020voxel2mesh} employs a GCN in the decoder to refine a sphere mesh to match a target shape.
As a more editable input, scene graphs are considered in \cite{dhamo2021graph} to generate and manipulate 3D scene meshes via a GCN-based variational auto-encoder. 


\noindent
\textbf{Human-related Mesh Reconstruction.}
For human body reconstruction, Kolotouros \textit{et al.}~\cite{kolotouros2019convolutional} improve the template-based mesh regression approach through a GCN to explicitly encode the template mesh structure and aggregate image features.
A bilayer graph is designed in \cite{yu2021joint} to represent the body shape and pose simultaneously, with a GCN deployed to jointly perform pose estimation and mesh regression. 
DC-GNet~\cite{zhou2021dc} also adopts a GCN and further takes into account occlusion cases where human bodies are incomplete in images.
The graph Transformer is exploited in Mesh Graphormer~\cite{lin2021mesh} for human mesh reconstruction. 
For reconstruction from 2D human pose, 
Pose2Mesh~\cite{choi2020pose2mesh} and GTRS~\cite{zheng2021lightweight} respectively builds on GCNs or graph Transformers for pose-guided mesh regression.
Besides human body, the face and hand reconstruction are also explored where GCN is a popular structure~\cite{lin2020towards, cheng2020faster, li2022interacting}. 


\vspace{1mm}
\noindent
\textbf{Discussion.}
Recently, the self-attention mechanism has been adopted for capturing global interactions~\cite{zhao2021point, lin2021mesh}. However, graph Transformers have not been adequately explored for general 3D object and scene reconstruction, and more graph Transformers are expected to be developed. In addition, in view of the large data size in 3D generation tasks, a practical direction would be promoting the efficiency of graph convolutional networks and graph Transformers. 

\vspace{-3mm}
\section{Medical Image Analysis}
\label{sec:medical}

The applications of GNNs on medical image analysis include, but are not limited to, brain activity investigation, disease diagnosis, and anatomy segmentation. 
\vspace{-0mm}

\subsection{Brain Activity Investigation}
\begin{figure}
    \centering
    \includegraphics[width=0.75\columnwidth]{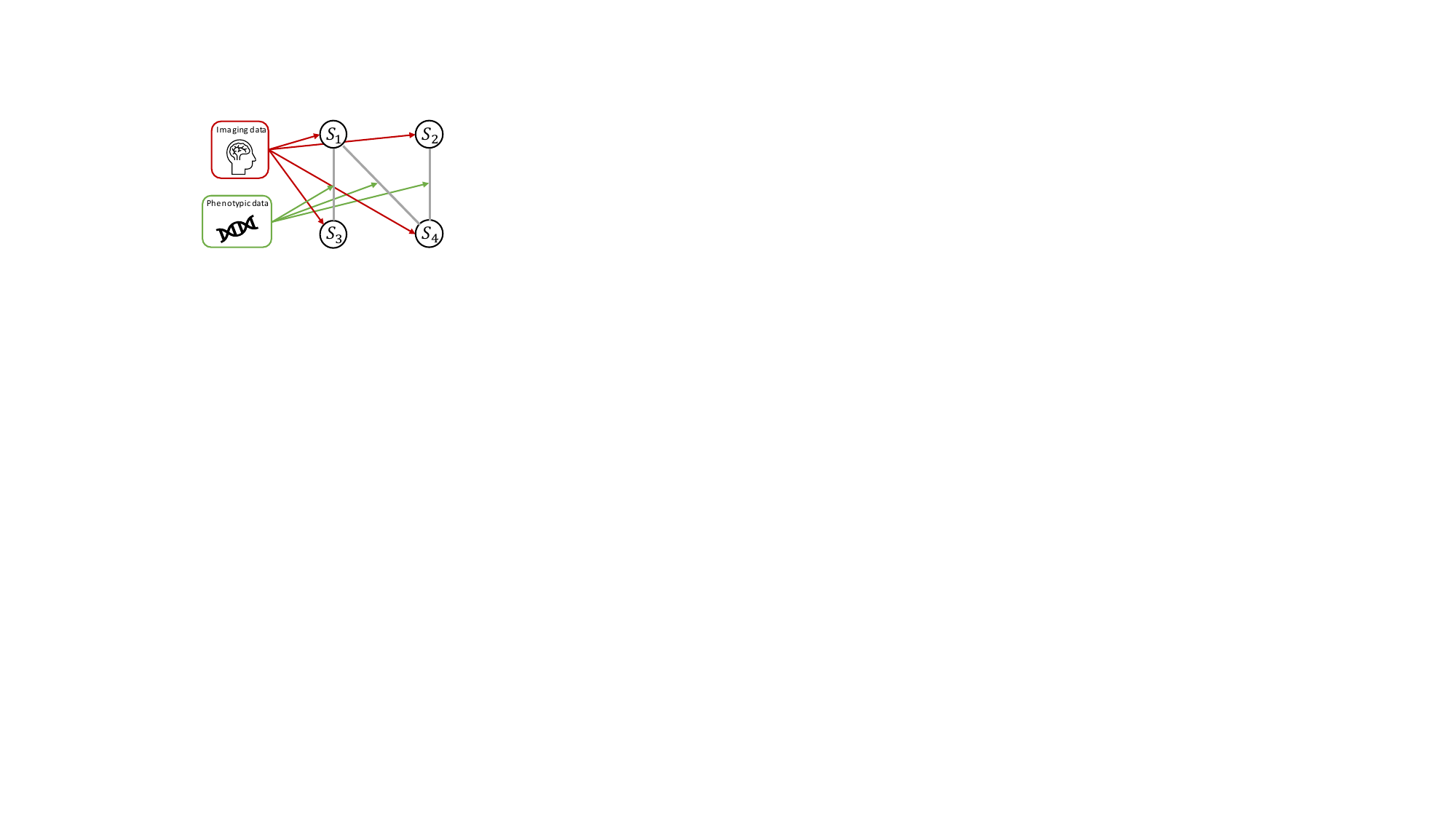}
    \caption{Subject-based brain analysis. $S_i$ denotes the $i$-th subject. Note that the node and edge features may vary depending on the task.}
    \label{fig:sub}
    \vspace{-0.3cm}
\end{figure}
\begin{figure}[!t]
	\centering
	\includegraphics[width=0.5\columnwidth]{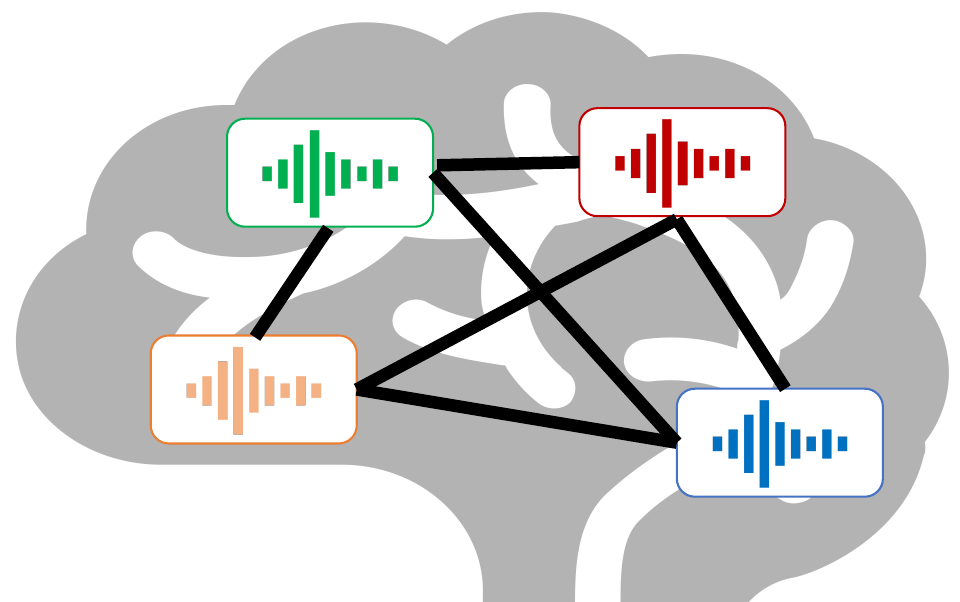}
	\caption{Region-based brain analysis. Colors denote brain regions with different functions/structures.\vspace{-2mm}}
	\label{fig:reg}
        \vspace{-2mm}
\end{figure}
Approaches for brain activity analysis can be roughly divided into two categories: subject- and region-based, according to the granularity of graph nodes. For subject-based methods (cf. Fig.~\ref{fig:sub}), each node comprises characteristics of a subject, and GNNs are applied to graphs consisting of $\hat{N}$ subjects:
\begin{align}
    \footnotesize
    \begin{split}
        S_i &= \Pi_{s}(\hat{I}_i),\\
        \mathcal{F}&=\mathtt{GNN}(\{S_i\}_{i=1}^{\hat{N}}),
    \label{eq:sub}
    \end{split}
\end{align}
where $\Pi_{s}(\cdot)$ denotes the graph construction process on top of $N$ subjects. $\hat{I}_i$ and $S_i$ represent the subject-level raw data and model input, respectively. $\mathcal{F}$ stands for the extracted graph representations. In comparison, region-based approaches construct graphs based on brain regions (cf. Fig.~\ref{fig:reg}), where each node contains the representation of a specific brain area. Similar to subject-based methodologies, region-based methods can be formulated as follows:
\begin{align}
    \footnotesize
    \begin{split}
        R_i&=\Pi_{r}(\tilde{I}_i),\\
        \mathcal{F}&=\mathtt{GNN}(\{R_i\}_{i=1}^{\tilde{N}}),
    \end{split}
    \label{eq:reg}
\end{align}
where $i=1,2,...,\tilde{N}$. $\Pi_{r}(\cdot)$ denotes the graph construction process on top of brain areas. $\tilde{I}_i$ and $R_i$ stand for the raw data and graph feature input of the $i$-th brain area, respectively.
\subsubsection{Brain Signal Disorder Recognition}
\noindent \textbf{Autism spectrum disorder (ASD).} Autism spectrum disorder is a type of developmental disorder that often adversely affects people's social communication and interaction. Recently, GNNs have been employed to identify ASD in functional magnetic resonance imaging (f-MRI)~\cite{heinsfeld2018identification}.

\textbf{Subject-based methods.} Parisot~\etal~\cite{parisot2017spectral} introduced spectral GCNs to conduct brain analysis in populations. 
The imaging feature of each subject serves as the vertex in the graph. 
The graph edge would be built if the similarity score of two subjects' imaging and non-imaging (e.g., phenotypic data) features is above an artificially defined threshold. The proposed GCN framework outperforms the archetypical ridge classifier by at least 5\% on different datasets. 
Anirudh~\etal~\cite{anirudh2019bootstrapping} addressed the instability problem in the step of population graph construction (with fewer edges) and introduced the bootstrap approach to construct an ensemble of multiple GCNs, which provides performance gains over~\cite{parisot2017spectral}. 
Rakhimberdina and Murata~\cite{rakhimberdina2019linear} replaced the spectral GCNs in~\cite{parisot2017spectral} with linear GCNs and optimized the edge construction step. 
Jiang~\etal~\cite{jiang2020hi} proposed hybrid GCNs that maintain a hierarchical architecture to jointly incorporate information from both subject and population levels. 
Peng~\etal~\cite{peng2022fedni} applied federated learning to train GCNs on inter-institutional data by first inpainting local networks with masked nodes and then training a global network across institutions.

\textbf{Region-based methods.} Ktena~\etal~\cite{ktena2017distance} used siamese GCNs to conduct distance metric learning on functional graphs of single subjects, outperforming the Principal Component Analysis (PCA) based baseline by over 10\% on different imaging sites. 
Based on the work~\cite{ktena2017distance}, Yao~\etal~\cite{yao2019triplet} proposed triplet GCNs that contrast matching and non-matching functional connectivity graphs at the same time. 
In addition, multi-scale templates~\cite{yao2019triplet} were employed to advance the graph representation learning results. 
Li~\etal~\cite{li2020pooling} modified the ranking-based pooling methods by adding the regularization, enabling better node selection and prediction interpretation. 
In BrainGNN~\cite{li2021braingnn}, ROI-aware graph convolution was introduced to encode the locality information into associated node embeddings. 
Kazi~\etal~\cite{kazi2019inceptiongcn} introduced inception GCN, which integrates spectral convolution with different kernel sizes to capture the intra- and inter-graph structural information.\\

\noindent \textbf{Other disorders.} Brain imaging can help distinguish patients with major depressive disorder (MDD) from normal ones. In~\cite{yao2020temporal}, Yao~\etal proposed an adaptive GCN framework to incorporate both spatial and temporal information from associated time-series f-MRI data. Yao~\etal~\cite{yao2021tensor} aligned multi-index representations of f-MRI and encoded them into a shared latent feature space for making MDD predictions. 
For the diagnosis of bipolar disorder, Yang~\etal~\cite{yang2019interpretable} employed GAT networks and hierarchical pooling strategies to deal with variable-sized inputs. 
Rakhimberdina and Murata~\cite{rakhimberdina2019linear} applied linear GCNs to tackle the diagnosis issue of schizophrenia that is a mental illness related to the barrier of perceiving the reality.
\vspace{-2mm}
\subsubsection{Brain State Decoding}
Zhang~\etal~\cite{zhang2021functional} used deep graph convolution to interpret Resting-State Functional MRI (rsf-MRI) data into associated cognitive task results spanning six cognitive domains. 
In~\cite{gadgil2020spatio}, the spatial-temporal graph convolution was employed to predict age and gender based on the BOLD signal of rsf-MRI data. 
Kim~\etal~\cite{kim2021learning} proposed to learn dynamic graph representations using spatial-temporal attention while Transformer serves as the encoder model to process a short sequence of brain connectome graphs. 

\subsection{Disease Diagnosis}
In this section, we mainly review the related literature on diagnosis of brain and chest diseases using GNNs. Specifically, for brain diseases, we involve Alzheimer’s and Parkinson’s diseases as two representatives. As for chest pathologies, we do not restrict the type of disease.
\vspace{-2mm}
\subsubsection{Brain Disease Identification}
\noindent \textbf{Alzheimer's disease (AD).} 
Alzheimer's disease is a type of neurodegenerative disease that gradually destroys brain cells. Methods for AD diagnosis are mostly based on subject-level information (cf. Fig.~\ref{fig:sub} and Eq.~\ref{eq:sub}).

Parisot~\etal~\cite{parisot2018disease} formulated the subject diagnosis issue as a graph labeling problem, where a spectral GCN was introduced to make diagnoses based on population analysis. As shown in Fig.~\ref{fig:sub}, imaging data provide node features, while phenotypic data are interpreted as edge weights. The proposed approach surpasses previous state-of-the-arts by about 5\% at least.
Zhao~\etal~\cite{zhao2019graphISBI} used a GCN-based approach to detect Mild cognitive impairment (MCI) in AD. In~\cite{liu2020identification}, Liu~\etal deployed a GCN model to identify early mild cognitive impairment (EMCI, an early stage of AD) with a multi-task feature selection method on both imaging and non-imaging data. Yu~\etal~\cite{yu2020multi} designed a multi-scale GCN framework that integrates each subject's structural information, functional information, and demographics (i.e., gender and age) for diagnosing EMCI based on population. 
Huang~\etal~\cite{huang2020edge} constructed an adaptive population graph with variational edges. The uncertainty of the adaptive graph was estimated using the proposed Monte-Carlo edge dropout~\cite{huang2020edge}. 
Ma~\etal~\cite{ma2020attention} incorporated local and global attention modules into GNNs. Besides, an attention-guided random walk strategy~\cite{ma2020attention} was applied to extract features from dynamic graphs (with variable-sized nodes). 
Wee~\etal~\cite{wee2019cortical} developed a cortical GNN model to take into account the cortical geometry and thickness information for identifying AD or MCI.\\

\noindent \textbf{Parkinson’s disease (PD).} Approaches for PD diagnosis mainly construct graphs based on brain regions (cf. Fig.~\ref{fig:reg} and Eq.~\ref{eq:reg}).
Zhang~\etal~\cite{zhang2018multi} implemented a multi-modal GCN model that fuses different brain image modalities for distinguishing PD from the healthy control, outperforming traditional diagnosis machine learning models by large margins. 
Kazi~\etal~\cite{kazi2019graph} developed a LSTM-based attention mechanism to learn subject-specific features by ranking multi-modal features, and the fused representation was used for making final decisions. 
In~\cite{zhang2020deep}, an end-to-end deep GCN framework was introduced to encode the cross-modality relationship into the graph while learning the mapping from structures to brain functions.
\vspace{-2mm}
\subsubsection{Chest Disease Analysis}
Chest disease analysis methods mainly build graphs upon anatomies or textual attributes, both of which are similar to the region-based schema (cf. Fig.~\ref{fig:reg} and Eq.~\ref{eq:reg}).
Yu~\etal~\cite{yu2021resgnet} treated each CT scan as a node and connected nodes that share highest similarities. Features extracted from CNNs are passed to a one-layer GCN to help capture the relationship between similar samples. 
Similarly, Wang~\etal~\cite{wang2021covid} proposed to fuse image-level and relation-aware features via a GCN. 
Here, the GCN model is used to gather neighborhood information for deciding whether the input CT scan has COVID-19 or not. 
Zhang~\etal~\cite{zhang2020radiology} built a knowledge graph of chest abnormality based on prior knowledge on chest findings. The obtained graph was integrated with a graph embedding module, which was appended to a off-shelf CNN feature extractor. The output features of the graph embedding module can be used for both disease classification and radiology report generation. 
Hou~\etal~\cite{hou2021multi} pre-trained the label co-occurrence graph on radiology reports and integrated the resulting label embeddings with image-level features. The fused representations are passed to the Transformer encoder and the following GCN layers for feature fusion. 
Liu~\etal~\cite{liu2020cross,liu2021act} introduced an anatomy-aware GCN model for mammogram detection. The anatomy-aware framework mainly involves two modules: i) a bipartite GCN to model the intrinsic geometrical and structural relations of ipsilateral views. ii) an inception GCN to model the structural similarities of bilateral views. The proposed anatomy-aware GCNs not only achieved state-of-the-art results on mammogram classification but also provided interpretable diagnosis results. 
Lian~\etal~\cite{lian2021structure} proposed a series of relation modules to capture the relations among thoracic diseases and anatomical structures. 
Chen~\etal~\cite{chen2022instance} proposed an instance importance-aware GCN for applying multi-instance learning to 3D medical diagnosis, which helps to learn complementary representations by exploiting importance- and feature-based topologies.
Zhao~\etal~\cite{zhao2021diagnose} proposed a general attribute-based medical image diagnosis framework by coupling probabilistic reasoning (Bayesian network) and neural reasoning (GCN) modules to model the causal relationships between attributes and diseases.

\subsection{Anatomy Segmentation}
Based on the segmentation targets, we roughly divide GNN-based segmentation approaches into three groups: brain surface segmentation, vessel segmentation, and other anatomical structure segmentation. Most of these segmentation approaches are built on top of certain structures, which can be regarded as extensions of the region-based approaches (cf. Fig.~\ref{fig:reg} and Eq.~\ref{eq:reg}).
\vspace{-2mm}
\subsubsection{Brain Surface Segmentation}
Cucurull~\etal~\cite{cucurull2018convolutional} exploited GCNs and GATs for mesh-based cerebral cortex segmentation. They found that either GCNs or GATs can obviously outperform traditional mesh segmentation approaches, sometimes by over 5\%. 
Gopinath~\etal~\cite{gopinath2019adaptive} used GCNs to learn spectral embeddings directly in a space defined by surface basis functions for brain surface parcellation.
Specifically, the proposed GCN model employs spectral filters to handle intrinsic surface representations, which were also validated in the task of brain parcellation.
To better handle impaired topology, Wu~\etal~\cite{wu2019intrinsic} proposed to directly conduct surface parcellation on top of the cortical manifold with the help of GCNs, which produced satisfactory results on surfaces that violate the spherical topology. 
He~\etal~\cite{he2020spectral} used the spectral graph Transformer to align multiple brain surfaces in the spectral domain, outperforming traditional GCN-based models by observable margins. 
\vspace{-2mm}
\subsubsection{Vessel Segmentation}
Wolterink~\etal~\cite{wolterink2019graph} used graph convolutional networks (GCNs) to forecast the spatial placement of vertices in a tubular surface mesh that divides the coronary artery lumen. Zhai~\etal~\cite{zhai2019linking} combined GCNs with CNNs to categorize pulmonary vascular trees into either arteries or veins. In practice, the hybrid vessel segmentation model is trained on constructed vessel graphs, where the label of each node is whether it belongs to the artery class or the vein. Shin~\etal~\cite{shin2019deep} proposed a unified model that consists of both CNNs and GNNs to leverage graphical connectivity for vessel segmentation. Specifically, the unified framework exploits local and global information from perspectives of appearances and structures, respectively. Noh~\etal~\cite{noh2020combining} extracted vessel connectivity matrices from fundus and fluorescein angiography images via a hierarchical GNN, which were used to assist the classification of retinal artery/vein classification. To tackle high variations in intracranial arteries, Chen~\etal~\cite{chen2020automated} introduced a GNN segmentation model and a hierarchical refinement module to integrate structural information with relational prior knowledge for the segmentation of intracranial arteries. Considering vessels in 3D images often have diverse sizes and shapes, Yao~\etal~\cite{yao2020graph} proposed a GCN-based point cloud network to incorporate the tubular prior knowledge into vessel segmentation, which not only helps to capture the global shape but also the local vascular structure. Yang~\etal~\cite{yang2020cpr} introduced a partial-residual GCN to take into account both position features of coronary arteries and associated imaging features to conduct automated anatomical segmentation. Zhang~\etal~\cite{zhang2021corlab} integrated the intra-artery relation with the inter-organ dependencies for capturing anatomical dependencies.
Zhao~\etal~\cite{zhao2022graph} proposed a cascaded deep neural network, where cross-network multi-scale feature fusion is performed between a CNN-based U-Net and a graph U-Net to effectively support high-quality vessel segmentation.
\vspace{-2mm}
\subsubsection{Segmentation of Other Anatomical Structures}
Selvan~\cite{selvan2018extraction} proposed to use a GNN-based auto-encoder to learn node features for airway extraction while a decoder was employed to predict edges between nodes. Garcia-Uceda Juarez~\etal~\cite{garcia2019joint} modified UNet~\cite{ronneberger2015u} by replacing the convolution layers in the bottleneck with graph convolution layers, which helps encode the node connectivity into latent embeddings. The resulting hybrid UNet outperformed the classic UNet in the task of airway segmentation. In~\cite{selvan2020graph}, a GNN-based graph refinement module was derived for to extract airway structures more accurately. Mukul~\etal~\cite{mukul2020uncertainty} formulated the post-processing procedure in pancreas and spleen segmentation as a semi-supervised graph labeling problem and proposed a GCN-based refinement strategy to replace the classic conditional random fields (CRFs) for segmentation mask post-processing. Tian~\etal~\cite{tian2020graph} applied GCNs to the interactive segmentation problem of prostate, where GCNs were responsible for refining the coarse contour produced by CNNs. Meng~\etal~\cite{meng2020cnn,meng2020regression} developed a hybrid U-shape model for optic cup/disc segmentation, where the encoder consists of convolution layers while the decoder comprises graph convolution layers. The encoder and decoder is connected via attention modules, which help the decoder to leverage the more location information from the convolutional encoder. Chao~\etal~\cite{chao2020lymph} presented a GNN that incorporates the global spatial priors of lymph node tumor into its learning process to further model the relations between lymph nodes. Yan~\etal~\cite{yan2019brain} first transform 3D MRI images into supervoxels, which were then passed to graph convolution layers to capture interconnections for brain tissue segmentation.
\vspace{-2mm}
\subsection{Discussion}
Graph representation learning is the core idea behind GNN-based brain signal analysis, while self-supervised learning has been shown to improve the generalization ability of representations in medical imaging~\cite{chen2019self,zhou2020comparing,zhou2021models,zhou2023preservation}. Therefore, it is promising to incorporate self-supervised learning into existing GNN-based analysis frameworks. For instance, neighboring nodes can be contrasted with non-neighboring ones to learn invariant and discriminative node features using noise-contrastive estimation~\cite{gutmann2010noise}.

Multi-disease graphs are necessary for capturing inter-disease relations to achieve reliable diagnoses. For example, a knowledge graph~\cite{ehrlinger2016towards,wang2017knowledge} can be integrated into a hierarchical multi-disease diagnosis system, where each node includes both imaging and non-imaging features of a specific disease. The knowledge graph can be built using domain knowledge or clinical descriptions associated with images (such as radiology reports~\cite{zhang2020contrastive,huang2021gloria,zhou2022generalized}). 

GNN-based approaches resort to building graphs for encoding relations into latent representations. In practice, these graphs can be incorporated into label-efficient learning to assist medical image segmentation with limited annotations. For instance, we can apply GNNs to aggregate the representations of neighboring classes with rich annotations to improve the representations of rare diseases. Moreover, conducting graph-based reasoning~\cite{wang2018zero} within GNN-based segmentation frameworks is also a promising direction, which helps improve the performance of segmentation models on unseen organs and diseases.


\section{Conclusions}
Despite the ground-breaking progress in perception, how to endow deep learning models with reasoning ability remains a formidable challenge for modern computer vision systems.
In this regard, GNN and graph Transformers have demonstrated significant flexibility and superiority in dealing with `relational' tasks.
To this end, we have presented the first comprehensive survey of GNN and graph Transformers in computer vision from a task-oriented perspective.
Specifically, a variety of classical and up-to-date algorithms are grouped into five categories according to the modality of input data, such as image, video, and point cloud.
By systematically sorting out the methodologies for each task,
we hope that this survey can shed light on more progress in the future.
By providing discussion regarding key innovations, limitations, and potential research directions,
we hope that readers can obtain new insights and go a further step towards human-like visual understanding.

\small
\bibliographystyle{IEEEtran}
\bibliography{IEEEabrv,main}

\end{document}